\documentclass[10pt,conference]{IEEEtran}
\IEEEoverridecommandlockouts

\usepackage{graphicx} 
\usepackage{tikz}
\usepackage{subcaption}
\usepackage{hyperref}
\usepackage{amssymb}
\usetikzlibrary{fit,backgrounds}
\usepackage{booktabs}
\usepackage{caption}
\usetikzlibrary{quantikz2, decorations.pathreplacing}
\def\mycdots{\ \scalebox{1.5}{$\cdots$}\ }
\def\BibTeX{{\rm B\kern-.05em{\sc i\kern-.025em b}\kern-.08em
    T\kern-.1667em\lower.7ex\hbox{E}\kern-.125emX}}

\title{Partitioned Hybrid Quantum Fourier Neural Operators for Scientific Quantum Machine Learning
}

\author{
  \IEEEauthorblockN{Paolo Marcandelli\IEEEauthorrefmark{1}, Yuanchun He\IEEEauthorrefmark{2}, Stefano Mariani\IEEEauthorrefmark{1},\\ Martina Siena\IEEEauthorrefmark{1},
  Stefano Markidis\IEEEauthorrefmark{2}}
  \IEEEauthorblockA{\IEEEauthorrefmark{1}Department of Civil and Environmental Engineering, Politecnico di Milano, Milan, Italy}
  \IEEEauthorblockA{\IEEEauthorrefmark{2}School of Electrical Engineering and Computer Science, KTH Royal Institute of technology, Stockholm, Sweden\\ Email: \texttt{paolo.marcandelli@polimi.it}}
}

\begin{document}

\maketitle

\begin{abstract}
We introduce the Partitioned Hybrid Quantum Fourier Neural Operator (PH-QFNO), a generalization of the Quantum Fourier Neural Operator (QFNO) for scientific machine learning. PH-QFNO partitions the Fourier operator computation across classical and quantum resources, enabling tunable quantum-classical hybridization and distributed execution across quantum and classical devices. The method extends QFNOs to higher dimensions and incorporates a message-passing framework to distribute data across different partitions. Input data are encoded into quantum states using unary encoding, and quantum circuit parameters are optimized using a variational scheme.  We implement PH-QFNO using PennyLane with PyTorch integration and evaluate it on Burgers’ equation, incompressible and compressible Navier-Stokes equations. We show that PH-QFNO recovers classical FNO accuracy. On incompressible Navier-Stokes, PH-QFNO achieves higher accuracy than its classical counterparts. Finally, we perform a sensitivity analysis under input noise, confirming improved stability of PH-QFNO over classical baselines.
\end{abstract}


\section{Introduction}
Quantum computing is emerging as a viable approach for solving problems across various domains, including cryptography, optimization, data analysis, and scientific computing~\cite{hegde2024beyond}. Within scientific computing, the numerical solution of partial differential equations (PDEs) remains a key application area. Classical numerical methods typically approach PDEs by discretizing them into systems of linear equations or eigenvalue problems. In contrast, quantum algorithms such as Harrow-Hassidim-Lloyd (HHL)~\cite{harrow2009quantum} and Quantum Phase Estimation (QPE)~\cite{nielsen2010quantum} provide fully quantum formulations for solving such problems, leveraging quantum principles including superposition, entanglement, and interference~\cite{markidis2024quantum}. However, these algorithms generally assume the availability of fault-tolerant quantum hardware, which remains beyond the capabilities of current Noisy Intermediate-Scale Quantum (NISQ) devices~\cite{preskill2018quantum}.

To overcome this limitation, hybrid quantum-classical approaches have been proposed~\cite{cerezo2021variational}. These approaches divide the computational pipeline between classical and quantum resources and typically rely on variational principles to optimize parameterized quantum circuits by minimizing an error or loss function~\cite{markidis2023programming}. Within this framework, quantum scientific machine learning~\cite{jaderberg2024potential} has been developed to approximate PDE solutions using quantum neural networks. These networks take collocation points as input and return approximate solutions of the underlying equations, with parameters trained using classical simulations, such as finite difference or finite element methods. This work adopts a hybrid variational framework to develop a solver for PDEs, targeting applications in Computational Fluid Dynamics (CFD), including Burgers’ and Navier-Stokes equations.

Quantum Fourier Neural Operators (QFNOs)~\cite{jain2023quantumfouriernetworkssolving} are a quantum analog of classical Fourier Neural Operators (FNOs)~\cite{FNO}, which learn operator mappings in spectral space by leveraging the Fourier transform. Many physical applications, such as fluid or electromagnetic problems, are efficiently represented in the spectral domain, where a limited number of Fourier modes often suffice and classical FNOs have been successfully deployed to solve such problems~\cite{costa2023deep}. QFNOs employ the Quantum Fourier Transform (QFT), which offers a polynomial complexity improvement over classical Fast Fourier Transforms~(FFTs). While QFNOs are promising, current formulations are limited. They lack a full-fledged implementation and generalization to higher dimensionality, and the ability to operate in distributed environments involving both classical and multiple quantum devices.

To address these limitations, we introduce a Partitioned Hybrid Quantum Fourier Neural Operator (PH-QFNO). The method partitions the QFNO computation across classical and quantum resources. Drawing inspiration from multidimensional FFTs, where multiple 1D FFTs are performed in parallel and can be further decomposed into smaller FFTs, we \emph{partition} the QFT-based operations across different quantum devices. PH-QFNO supports adjustable levels of quantum-classical computation, from fully classical FNO to fully quantum QFNO. The contributions of this work are as follows:
\begin{itemize}
    \item A formulation and methodology for combining classical and quantum computation within QFNOs, extended to 2D and 3D problems. The implementation leverages PennyLane and its integration with PyTorch for CFD applications.
    \item A distributed execution framework using MPI~\cite{gropp1999using} to assign partitions to CPU cores and QPUs. Quantum resources are simulated with PennyLane’s noiseless state vector simulator~\cite{bergholm2018pennylane}, supporting different degrees of hybridization.
    \item Evaluation of PH-QFNO on Burgers’ and Navier-Stokes equations. Across hybrid configurations, PH-QFNO consistently outperforms classical FNO in terms of accuracy.
    \item Analysis of robustness under input noise. Fidelity-based assessments across noise levels show that PH-QFNO is more stable than its classical counterparts.
\end{itemize}

\section{Theoretical Background}
This section provides the theoretical background underpinning the development of our PH-QFNO for both 1D and 2D domains.
\subsection{Learning Operators and Neural Operators}\label{learningoper}
Learning operators are designed to learn the mapping between function spaces from a finite collection of observed input-output pairs, see Refs.~\cite{LearningOperator}~and~\cite{FNO} for more details. Neural Operators (NOs) represent their multi-layer neural implementation, and we will apply these operators to the specific case of PDEs. Since we work with numerical data, we only have access to discrete representations of the domain. Then, the discretization-invariance property of NOs will be crucial in this setting.

Let $D\subset\mathbb{R}^d$ be the spatial domain and $D_j \subset D$ its discretized version, and let $\mathcal{A}=\mathcal{A}(D_j;\mathbb{R}^{d_a})$ and $\mathcal{U}=\mathcal{U}(D_j;\mathbb{R}^{d_u})$ be separable Banach spaces of functions taking values in $\mathbb{R}^{d_a}$ and $\mathbb{R}^{d_u}$ respectively. Then, let $\mathcal{G}^\dagger: \mathcal{A}\to\mathcal{U}$ be a typically non-linear map that we aim to approximate. 
The NO proposed in Ref.~\cite{LearningOperator} uses a point-wise function $P$ to map the input $\{a\,:\, D_j\to\mathbb{R}^{d_a}\}$ to a higher dimensional space $ \{v_0\,:\, D_j'\to\mathbb{R}^{d_{v_0}}\}$. Then, the operator is formulated as an iterative architecture $v_0\to v_1\to\cdots\to v_{T}$ where $v_j\,:\,D_j'\to\mathbb{R}^{d_v}$ for $j=0,1,\cdots,T-1$.  In the end, using a point-wise function $Q$ the last hidden representation  $\{v_T\,:\, D_j'\to\mathbb{R}^{d_v}\}$ is mapped to the output function $ \{u\,:\, D_j\to\mathbb{R}^{d_{u}}\}$. In our paper we will study the NO applied to PDEs. We seek to learn the mapping $\mathcal{G}^\dagger\,:\,a\to u$, where $a\in\mathcal{L}^2(D_j;\mathbb{R})$ is the initial condition, and $u(t)\in\mathcal{H}^s(D_j;\mathbb{R})$ is the solution at later time $t$. We set $T=1$ to reduce the NO at one single layer due to computational costs reasons.

\subsection{Unary Encoding}\label{unaryencoding}
The initial phase of the quantum circuit entails embedding a matrix, in 2D or 3D, into the unary basis. Accordingly, we provide a concise exposition of the underlying mechanism by which this embedding is effected in the elementary one-dimensional scenario. 

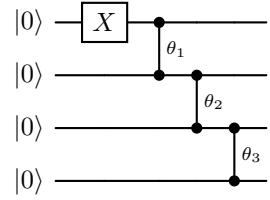
\begin{figure}[htbp]
\centering
\begin{quantikz}[column sep=10pt, row sep={20pt,between origins}, control/.append style={minimum size=3mm}]
\lstick{\ket{0}} &\gate{X}&\ctrl[wire style={"\theta_1"}]{1}&&&\\
\lstick{\ket{0}} &&\control{}&\ctrl[wire style={"\theta_2"}]{1}&&\\
\lstick{\ket{0}} &&&\control{}&\ctrl[wire style={"\theta_3"}]{1}&\\
\lstick{\ket{0}} &&&&\control{}&
\end{quantikz}
\caption{Quantum circuit for one-dimensional encoding with four qubits. An $X$ gate is applied to the first wire and $RBS(\theta)$ gates to the sequential couples of qubits. The scheme can be generalized to n qubits.}
\label{encode}
\end{figure}

We begin with a one-dimensional vector $[x_0, \dots, x_{n-1}]$, which is normalized such that the last element is given by $x_{n-1} = \sqrt{1 - \sum_{i=0}^{n-2} x_i^2}$.
Next, we encode the normalized vector using the quantum circuit shown in Fig.~\ref{encode}, taken from  \cite{Cherrat2024quantumvision}, \cite{Johri2021} and \cite{quantumdeepONet}. 

We start with the state \(\ket{0\cdots 0}\) composed of \(n\) qubits. An \(X\) gate is applied to the first qubit, resulting in the state \(\ket{10\cdots 0}\) (following the Pennylane convention for qubit ordering). Subsequently, we apply the \(RBS\) (Reconfigurable Beam Splitter) gate on subsequent couples of qubits:
\begin{equation}
    RBS(\theta) = \begin{pmatrix}
    1 & 0 & 0 & 0 \\
    0 & \cos(\theta) & \sin(\theta) & 0 \\
    0 & -\sin(\theta) & \cos(\theta) & 0 \\
    0 & 0 & 0 & 1 
    \end{pmatrix}.
    \label{RBSgate}
\end{equation}
This gate performs the rotation
$\ket{10} \rightarrow \cos(\theta)\ket{10} + \sin(\theta)\ket{01}, \quad \ket{01} \rightarrow \cos(\theta)\ket{01} - \sin(\theta)\ket{10}$,
on the qubit couple, leaving the remaining basis states unchanged.
To encode a vector in $\mathbb{R}^n$, $n-1$ parameters $\theta$ are required, which can be recursively defined as


\begin{equation}\label{eq1}
    \theta_k = \arccos\left( \frac{x_k}{\prod_{j=0}^{k-1} \sin(\theta_j)} \right), \quad \text{for } k = 1, \dots, n-2
\end{equation}
where $\theta_0 = \arccos(x_0)$.
In the end, we obtain the following state:
\begin{align}
    \ket{\psi} =\ & \cos(\theta_0)\ket{10\cdots0} + \cos(\theta_1)\sin(\theta_0)\ket{01\cdots0} + \nonumber\\
    & \cdots + \sin(\theta_{n-2})\cdots\sin(\theta_0)\ket{00\cdots1}.
\end{align}

Substituting the definitions from Eq.~\ref{eq1} into the expression above, we obtain

\begin{align}
    \ket{\psi} = x_0 \ket{10\cdots0} + x_1\ket{01\cdots0} + \cdots + x_{n-1}\ket{00\cdots1}.
\end{align}

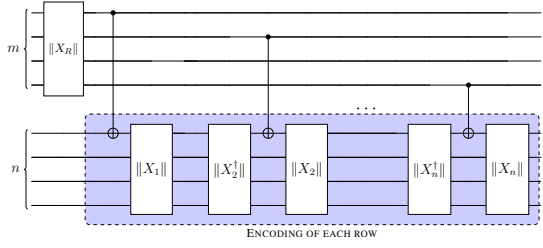
\begin{figure}[htb]
\centering
\resizebox{0.4\textwidth}{!}{

\begin{quantikz}[column sep=10pt, row sep={20pt,between origins}, phase/.append style={scale=2}, circlewc/.append style={scale=2}]
\lstick[4]{\textsc{\large $m$}} & \gate[4]{\|X_R\|} & \qw & \ctrl{5}  & \qw & \qw & \qw & \qw  & \qw  & \qw  & \qw & \qw  & \qw & \qw & \qw & \qw &   \\
                & \qw & \qw & \qw & \qw  & \qw & \qw & \qw & \ctrl{4} & \qw & \qw  & \qw & \qw & \qw & \qw & \qw & \\
                & \qw & \qw & \qw & \qw & & \qw \qw & \qw & \qw & \qw & \qw     & \qw & \qw & \qw & \qw & \qw  &\\
                & \qw & \qw & \qw & \qw  & \qw  & \qw      & \qw   & \qw & \qw & \qw & \qw  & \qw & \qw & \ctrl{2}  & \qw & \qw \\
               & \setwiretype{n} & \setwiretype{n} & \setwiretype{n} & \setwiretype{n} & \setwiretype{n} & \setwiretype{n} & \setwiretype{n} & \setwiretype{n}  & \setwiretype{n} & \setwiretype{n} & \mycdots\\
\lstick[4]{\textsc{\large $n$}} 
  & & \gategroup[4,steps=14,style={dashed,rounded
corners,fill=blue!20, inner
sep=5pt},background,label style={label
position=below,anchor=north,yshift=-0.2cm}]{{\sc
Encoding of each row}} & \targ{} 
  & \gate[4,disable auto height]{\textsc{\large $\|X_{1}\|$}} 
  & \qw & 
  & \gate[4,disable auto height]{\textsc{\large $\|X_{2}^\dagger\|$}}
  & \targ{} 
  & \gate[4,disable auto height]{\textsc{\large $\|X_{2}\|$}} 
  & \qw  & \qw  & \qw &   \gate[4,disable auto height]{\textsc{\large $\|X_{n}^\dagger\|$}}  & \targ{} & \gate[4,disable auto height]{\textsc{\large $\|X_{n}\|$}} & \\
                & \qw  & \qw  & \qw & \qw & \qw & \qw & \qw & \qw & \qw  & \qw  & \qw & \qw  & \qw & \qw & \qw &\\
                & \qw  & \qw & \qw & \qw & \qw & \qw & \qw & \qw & \qw& \qw & \qw &  \qw  & \qw & \qw & \qw &  \\
                & \qw  & \qw & \qw & \qw & \qw & \qw & \qw & \qw & \qw& \qw  & \qw & \qw & \qw  & \qw & \qw & 
\end{quantikz}}
\caption{Two-dimensional unary encoding for matrix $X\in \mathbb{R}^{m\times n}$. The initial state vector is given by the two-register state $\ket{0}^{\otimes m} \otimes \ket{0}^{\otimes n}$. Each block represents the one-dimensional encoding. In the first register the norm of each row $\|X_R\|$ is encoded, while in the second one  we sequentially encode and decode the values of each row. The result is the encoding of the normalized matrix $X$.}
\label{encodediagram}
\end{figure}

The method can be straightforwardly generalized to higher-dimensional spaces, such as $\mathbb{R}^{m\times n}$ or $\mathbb{R}^{m\times n\times k}$. For example, to encode a 2D matrix $X \in \mathbb{R}^{m\times n}$, a fundamental prerequisite for the development of novel 3D circuits, we follow the procedure illustrated in Fig.~\ref{encodediagram}.

First, we normalize the matrix and encode the norm of each row into the first register using the previously described method. This yields the state:
\begin{equation*}
    \ket{\psi} = \frac{1}{\|X\|}\left(\|X_1\| \underbrace{\ket{e_1}}_{m \text{\, dim}} \overbrace{\ket{e_0}}^{n \text{\, dim}} + \cdots+\|X_m\|\ket{e_m}\ket{e_0} \right)
\end{equation*}
where $\ket{e_i}$ is the Hamming weight one basis state with $1$ in position $i$, $\ket{e_0} = \ket{\bar{0}}$ and \(\|X_i\|\) denote the norm of row \(i\). 

Next, we proceed by encoding the values of each normalized row into the second register. The mathematical proof is straightforward when one follows the structure depicted in Fig.~\ref{encodediagram} and employs the unitarity of the $RBS(\theta)$ gate. In the end, we obtain the desired state: 
\begin{align*}
        \ket{X} = \frac{1}{\|X\|}\left(\sum_{i,j=1}^{m,n}x_{ij}\ket{e_i}\ket{e_j}\right).
\end{align*}
 
It is worth noting that the unary (one-hot) encoding strategy incurs only \emph{linear} resource overhead. Specifically, to embed a classical vector of dimension \(d\) one requires \(d\) qubits and exactly \(d-1\) two-mode RBS gates, yielding a circuit depth of \(O(d)\). No high-order multipartite entanglement is generated beyond the pairwise amplitude transfers, which greatly simplifies both state preparation and measurement.

By contrast, other encoding methods impose significantly greater computational and circuit complexity. For example, amplitude encoding achieves a compact footprint of \(\lceil\log_2 d\rceil\) qubits but typically demands \(O(d\log d)\) (or worse) gate depth, relies on numerous multi-controlled rotations, and produces complex entanglement patterns that increases error accumulation and experimental overhead.

Thus, although unary encoding trades qubit count for gate simplicity, it uniquely enables the preparation of arbitrary one-hot superpositions with only nearest-neighbor operations, avoiding any exponential blow-up in gate or entanglement complexity that plagues alternatives such as amplitude encoding.

\subsection{Quantum Fourier Transform }\label{unaryqft}
In this subsection, we detail the implementation of the Quantum Fourier Transform (QFT) in the unary basis, following the approach in~\cite{jain2023quantumfouriernetworkssolving}. The algorithm is a quantum adaptation of the classical Cooley–Tukey method~\cite{Cooley,markidis2024quantum}, which efficiently computes the FFT of a vector $x\in\mathbb{R}^n$.

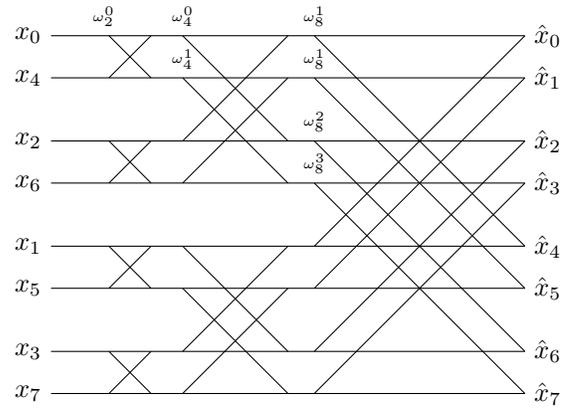
\begin{figure}[htb]
\centering

    \begin{tikzpicture}[x=0.7cm, y=0.7cm]

        \node[left] at (0,0) {$x_0$};
        \node[above] at (1,0) {\tiny$\omega_2^0$};
        \node[above] at (2.5,0) {\tiny$\omega_4^0$};
        \node[above] at (2.5,-0.8) {\tiny$\omega_4^1$};
        \node[above] at (5,0) {\tiny$\omega_8^1$};
        \node[above] at (5,-0.8) {\tiny$\omega_8^1$};
        \node[right] at (9,0) {$\hat{x}_0$};
        \draw (0,0) -- (9,0);
        \node[left] at (0,-0.8) {$x_4$};
        \node[right] at (9,-0.8) {$\hat{x}_1$};
        \draw (0,-0.8) -- (9,-0.8);
        
        \node[left] at (0,-2) {$x_2$};
        \node[right] at (9,-2) {$\hat{x}_2$};
        \node[above] at (5,-2) {\tiny$\omega_8^2$};
        \node[above] at (5,-2.8) {\tiny$\omega_8^3$};
        \draw (0,-2) -- (9,-2);
        \node[left] at (0,-2.8) {$x_6$};
        \node[right] at (9,-2.8) {$\hat{x}_3$};
        \draw (0,-2.8) -- (9,-2.8);
        
        \node[left] at (0,-4) {$x_1$};
        \node[right] at (9,-4) {$\hat{x}_4$};
        \draw (0,-4) -- (9,-4);
        \node[left] at (0,-4.8) {$x_5$};
        \node[right] at (9,-4.8) {$\hat{x}_5$};
        \draw (0,-4.8) -- (9,-4.8);
        
        \node[left] at (0,-6) {$x_3$};
        \node[right] at (9,-6) {$\hat{x}_6$};
        \draw (0,-6) -- (9,-6);
        \node[left] at (0,-6.8) {$x_7$};
        \node[right] at (9,-6.8) {$\hat{x}_7$};
        \draw (0,-6.8) -- (9,-6.8);
        
        \def\crosssize{0.4}
        
        \coordinate (C1) at (1.5,-0.4);
        \draw ($(C1)+(-\crosssize,-\crosssize)$) -- ($(C1)+(\crosssize,\crosssize)$);
        \draw ($(C1)+(-\crosssize,\crosssize)$) -- ($(C1)+(\crosssize,-\crosssize)$);
        
        \coordinate (C2) at (1.5,-2.4);
        \draw ($(C2)+(-\crosssize,-\crosssize)$) -- ($(C2)+(\crosssize,\crosssize)$);
        \draw ($(C2)+(-\crosssize,\crosssize)$) -- ($(C2)+(\crosssize,-\crosssize)$);
        
        \coordinate (C3) at (1.5,-4.4);
        \draw ($(C3)+(-\crosssize,-\crosssize)$) -- ($(C3)+(\crosssize,\crosssize)$);
        \draw ($(C3)+(-\crosssize,\crosssize)$) -- ($(C3)+(\crosssize,-\crosssize)$);
        
        \coordinate (C4) at (1.5,-6.4);
        \draw ($(C4)+(-\crosssize,-\crosssize)$) -- ($(C4)+(\crosssize,\crosssize)$);
        \draw ($(C4)+(-\crosssize,\crosssize)$) -- ($(C4)+(\crosssize,-\crosssize)$);
        
        \def\crosssizes{1}
        
        \coordinate (C5) at (3.5,-1);
        \draw ($(C5)+(-\crosssizes,-\crosssizes)$) -- ($(C5)+(\crosssizes,\crosssizes)$);
        \draw ($(C5)+(-\crosssizes,\crosssizes)$) -- ($(C5)+(\crosssizes,-\crosssizes)$);
        
        \coordinate (C6) at (3.5,-1.8);
        \draw ($(C6)+(-\crosssizes,-\crosssizes)$) -- ($(C6)+(\crosssizes,\crosssizes)$);
        \draw ($(C6)+(-\crosssizes,\crosssizes)$) -- ($(C6)+(\crosssizes,-\crosssizes)$);
        
        \coordinate (C7) at (3.5,-5);
        \draw ($(C7)+(-\crosssizes,-\crosssizes)$) -- ($(C7)+(\crosssizes,\crosssizes)$);
        \draw ($(C7)+(-\crosssizes,\crosssizes)$) -- ($(C7)+(\crosssizes,-\crosssizes)$);
        
        \coordinate (C8) at (3.5,-5.8);
        \draw ($(C8)+(-\crosssizes,-\crosssizes)$) -- ($(C8)+(\crosssizes,\crosssizes)$);
        \draw ($(C8)+(-\crosssizes,\crosssizes)$) -- ($(C8)+(\crosssizes,-\crosssizes)$);
        
        \def\crosssizess{2}
        \coordinate (C9) at (7,-2);
        \draw ($(C9)+(-\crosssizess,-\crosssizess)$) -- ($(C9)+(\crosssizess,\crosssizess)$);
        \draw ($(C9)+(-\crosssizess,\crosssizess)$) -- ($(C9)+(\crosssizess,-\crosssizess)$);
        
        \coordinate (C10) at (7,-2.8);
        \draw ($(C10)+(-\crosssizess,-\crosssizess)$) -- ($(C10)+(\crosssizess,\crosssizess)$);
        \draw ($(C10)+(-\crosssizess,\crosssizess)$) -- ($(C10)+(\crosssizess,-\crosssizess)$);
        
        \coordinate (C11) at (7,-4);
        \draw ($(C11)+(-\crosssizess,-\crosssizess)$) -- ($(C11)+(\crosssizess,\crosssizess)$);
        \draw ($(C11)+(-\crosssizess,\crosssizess)$) -- ($(C11)+(\crosssizess,-\crosssizess)$);
        
        \coordinate (C12) at (7,-4.8);
        \draw ($(C12)+(-\crosssizess,-\crosssizess)$) -- ($(C12)+(\crosssizess,\crosssizess)$);
        \draw ($(C12)+(-\crosssizess,\crosssizess)$) -- ($(C12)+(\crosssizess,-\crosssizess)$);

    \end{tikzpicture}
    \caption{Butterfly diagram of the Cooley-Tukey algorithm, which performs the FFT of the permuted input vector $x\in \mathbb{R}^8$. Each cross performs an elementary radix-2 operation with the root of unity $\omega_n^k=e^{i2\pi k/n}$.}
    \label{CooleyTukey}
    
\end{figure}

\noindent \textbf{Classical Cooley-Tukey algorithm:} 
The procedure begins by permuting the indices of the vector $x\in\mathbb{R}^n$ according to a bit-reversal scheme. The core computational step is known as the butterfly-shaped cross operation, see Fig.~\ref{CooleyTukey}.  

In this diagram, each cross 
\begin{center}
\begin{tikzpicture}

    \draw (0,0) -- (3,0);
    \node[above] at (1,0) {$\omega_n^k$};
    \node[left] at (0,0) {$a$};
    \node[right] at (3,0) {$a+\omega_n^kb$};
    \draw (0,-0.8) -- (3,-0.8);
    \node[left] at (0,-0.8) {$b$};
    \node[right] at (3,-0.8) {$a-\omega_n^kb$};
    \def\crosssize{0.4}
    
    \coordinate (C1) at (1.5,-0.4);
    \draw ($(C1)+(-\crosssize,-\crosssize)$) -- ($(C1)+(\crosssize,\crosssize)$);
    \draw ($(C1)+(-\crosssize,\crosssize)$) -- ($(C1)+(\crosssize,-\crosssize)$);
\end{tikzpicture}
\end{center}

is applied iteratively to pairs of input elements and maps the input pair $[a,b]$ as follows:
\[
\begin{pmatrix}
    1&\omega_n^k\\
    1&-\omega_n^k
\end{pmatrix}\begin{pmatrix} a \\ b \end{pmatrix} =
\begin{pmatrix} a + \omega_n^k\,b \\ a - \omega_n^k\,b \end{pmatrix},
\]
where $\omega_n^k = e^{i\frac{2\pi}{n}k},$ \(n=2,4,8\) denotes the number of inputs at each stage, and $k \in \{0,1,\ldots,\tfrac{n}{2}-1\}$. 

\begin{figure}[htb]
\centering
\resizebox{0.4\textwidth}{!}{
\begin{quantikz}[column sep=10pt, row sep={20pt,between origins}]
   &\gate[1]{\omega_2^0}&\ctrl[ ]{1}&\gate[1]{ \omega_4^0}&\ctrl[ ]{2}&&\gate[1]{ \omega_8^0}&\ctrl[ ]{4}&&&&\\
   &&\control{}&\gate[1]{\omega_4^1}&&\ctrl[ , wire label style={above}]{2}&\gate[1]{ \omega_8^1}&&\ctrl[ ]{4}&&&\\
   &\gate[1]{ \omega_2^0}&\ctrl[ ]{1}&&\control{}&&\gate[1]{ \omega_8^2}&&&\ctrl[ ]{4}&&\\
   &&\control{}&&&\control{}&\gate[1]{ \omega_8^3}&&&&\ctrl[ ]{4}&\\
   &\gate[1]{ \omega_2^0}&\ctrl[ ]{1}&\gate[1]{ \omega_4^0}&\ctrl[ ]{2}&&&\control{}&&&&\\
   &&\control{}&\gate[1]{ \omega_4^1}&&\ctrl[ ]{2}&&&\control{}&&&\\
   &\gate[1]{ \omega_2^0}&\ctrl[ ]{1}&&\control{}&&&&&\control{}&&\\
   &&\control{}&&&\control{}&&&&&\control{}&
\end{quantikz}
}
\caption{Quantum circuit designed to carry out the Fourier Transform in the unary basis. Single-qubit operations are phase gates, while vertical connections represent RBS gates with a rotation angle of $\pi/4$. This arrangement mirrors the classical FFT butterfly structure by replacing each radix-2 operation with a phase gate followed by an RBS gate. The amplitudes in the input state vector $\ket{x}=\sum_{i=0}^7x_i\ket{e_i}$ properly permuted and encoded in the unary basis, are transformed into the desired output state.}
\label{QFT}
\end{figure}
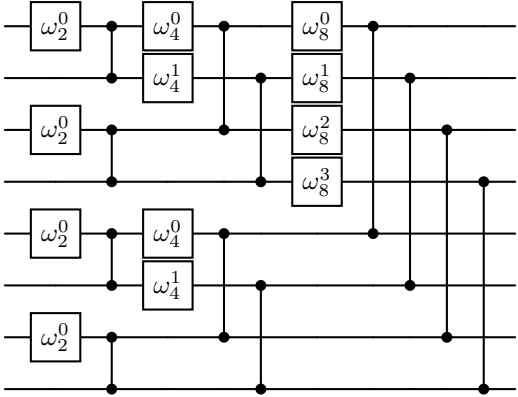

After these operations, the output is the desired Fourier transformed vector:
$$\hat{x} = [\hat{x}_0, \hat{x}_1, \hat{x}_2, \hat{x}_3, \hat{x}_4, \hat{x}_5, \hat{x}_6, \hat{x}_7].$$
\\
\noindent \textbf{Unary Quantum Fourier Transform:} Even in the quantum case, the indices of the vector to be transformed must be correctly permuted before applying the QFT. Then, the amplitudes $x_i\in x$ in the state vector should be encoded with the correct permutation, here indicated by the $\lfloor\,\,\rfloor$ symbol:
\begin{equation}\label{qftstate5}
    \ket{x} = \sum_{i=0}^{7} x_{\lfloor i\rfloor}\,\ket{e_i}
\end{equation}
Then, we apply the quantum analog of the Cooley-Tukey algorithm as described in \cite{jain2023quantumfouriernetworkssolving}. By following the Pennylane convention of ordering of qubits, the classical cross operation can be obtained by the following quantum diagram
\begin{center}
\begin{quantikz}
    \lstick[2]{\small $a\ket{10}+b\ket{01}$} & \gate[1]{\omega_n^k} & \ctrl{1} &\rstick[2]{\small$\frac{1}{\sqrt{2}}[(a+\omega_n^k\,b)\ket{10}$ \\ $+ (a-\omega_n^k\,b)\ket{01}],$}  \\
    & \qw & \control{} & 
\end{quantikz}
\end{center}
which is equivalent to the unitary operator
\begin{equation}
    U_{QFT} = \begin{pmatrix}
        1 & 0 & 0 & 0\\[1mm]
        0 & \frac{\omega_n^k}{\sqrt{2}} & \frac{1}{\sqrt{2}} & 0\\[1mm]
        0 & -\frac{\omega_n^k}{\sqrt{2}} & \frac{1}{\sqrt{2}} & 0\\[1mm]
        0 & 0 & 0 & -\omega_n^k
    \end{pmatrix},
\end{equation}
and acts on the state \(a\ket{10}+b\ket{01}\) as
\begin{equation*}
    U_{QFT}(a\ket{10}+b\ket{01}) = \frac{1}{\sqrt{2}}\Big[(a+\omega_n^k\,b)\ket{10} + (a-\omega_n^k\,b)\ket{01}\Big].
\end{equation*}
By applying the quantum butterfly circuit as depicted in Fig.~\ref{QFT}, one recovers the desired transformation as in the classical framework.

Applying Fourier Transform to both two and three dimensional matrices follows immediately the operation for simple vector $x\in\mathbb{R}^n$. For two-register circuit \cite{Johri2021}, before encoding, the matrix $X\in\mathbb{R}^{m\times n}$ must be subjected to a column permutation as previously described:
\begin{equation}\label{qftstate1}
    \ket{X} = \frac{1}{\|X\|}\left(\sum_{i,j=1}^{m,n} x_{i\lfloor j\rfloor}\,\ket{e_i}\ket{e_j}\right).
\end{equation}
In this way, one can apply the $QFT$ operator $\hat{Q}_{FT}$ to the second register and get:
\begin{equation}
    \begin{gathered}\label{qftstate2}
        \frac{1}{\|X\|}(\mathbb{I}\otimes\hat{Q}_{FT})\left(\sum_{i,j=1}^{m,n} x_{i\lfloor j\rfloor}\,\ket{e_i}\ket{e_j}\right)=\\
        \frac{1}{\|X\|}\left(\sum_{i,j=1}^{m,n} \hat{x}_{i j}\,\ket{e_i}\ket{e_j}\right).
    \end{gathered}
\end{equation}
By doing so, we reproduce the one dimensional $QFT$ of each row of the matrix $X$ preserving the unary basis.\\
The three register circuit QFT is described in the next section.

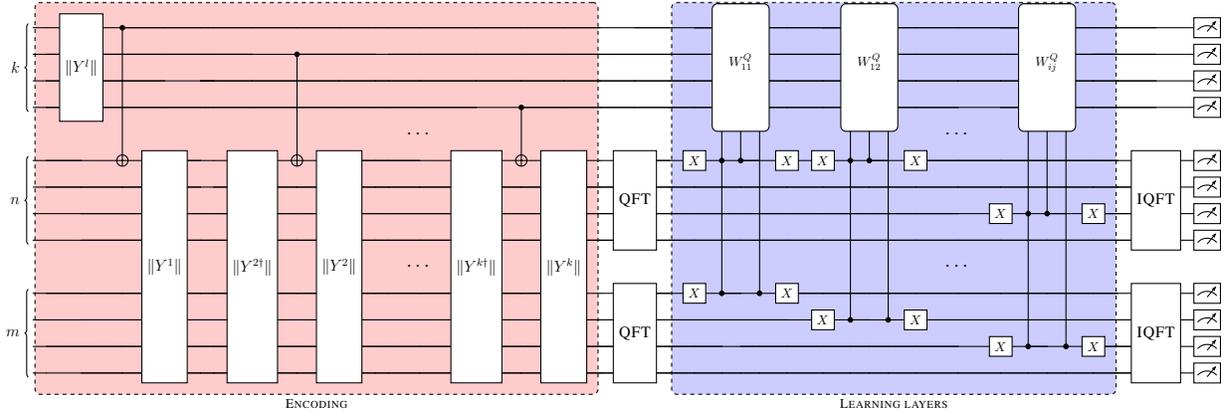
\begin{figure*}
\centering
\resizebox{0.9\textwidth}{!}{
\begin{quantikz}[column sep=10pt, row sep={20pt,between origins}]
\lstick[4]{\textsc{\large $k$}} & \gategroup[14,steps=15,style={dashed,rounded
corners,fill=red!20, inner
sep=5pt},background,label style={label
position=below,anchor=north,yshift=-0.2cm}]{{\sc
Encoding}}&\gate[4]{\textsc{\large $\|Y^l\|$}} & \ctrl{5}  & \qw & \qw & \qw & \qw  & \qw  & \qw  & \qw & \qw  & \qw & \qw & \qw & \qw & \qw  && \qw & \gategroup[14,steps=16,style={dashed,rounded
corners,fill=blue!20, inner
sep=5pt},background,label style={label
position=below,anchor=north,yshift=-0.2cm}]{{\sc
Learning layers}} & \qw\gategroup[4,steps=3,style={rounded corners, fill=white!, inner xsep=2pt},label style={label
position=below,anchor=north,yshift= 2cm}]{{\sc
$W^Q_{11}$}}\qw  & \qw & \qw & \qw & \qw & \qw\gategroup[4,steps=3,style={rounded corners, fill=white!, inner xsep=2pt},label style={label
position=below,anchor=north,yshift= 2cm}]{{\sc
$W^Q_{12}$}}\qw  & \qw  & \qw & \qw & \qw & \qw & \qw\gategroup[4,steps=3,style={rounded corners, fill=white!, inner xsep=2pt},label style={label
position=below,anchor=north,yshift= 2cm}]{{\sc
$W^Q_{ij}$}}\qw   & \qw  & \qw  & \qw & \qw & \qw  & \meter{} \\
                && \qw & \qw & \qw  & \qw & \qw & \qw & \ctrl{4} & \qw & \qw  && \qw & \qw & \qw & \qw & \qw & \qw & \qw  &  \qw & \qw & \qw & \qw & \qw & \qw & \qw & \qw  & \qw & \qw & \qw  & \qw  & \qw & \qw & \qw  & \qw & \qw  \qw & \qw  & \meter{}\\
                && \qw & \qw & \qw & & \qw \qw & \qw & \qw & \qw & \qw     & \qw & \qw & \qw & \qw & \qw  & \qw & \qw & \qw & \qw &  \qw  & \qw & \qw & \qw & \qw & \qw && \qw & \qw & \qw  & \qw & \qw & \qw & \qw  & \qw & \qw & \qw  & \meter{}\\
                && \qw & \qw & \qw  & \qw  & \qw      & \qw   & \qw & \qw & \qw & \qw  & \qw & \qw & \ctrl{2}  & \qw & \qw & \qw  & \qw & \qw & \qw & \qw & \qw & \qw   & \qw & \qw & \qw & \qw & \qw  & \qw && \qw & \qw & \qw & \qw & \qw & \qw &  \meter{}\\
               & \setwiretype{n} & \setwiretype{n} & \setwiretype{n}& \setwiretype{n} & \setwiretype{n} & \setwiretype{n} & \setwiretype{n} & \setwiretype{n} & \setwiretype{n}  & \setwiretype{n} & \setwiretype{n} & \mycdots \setwiretype{n} & \setwiretype{n}  & \setwiretype{n}  & \setwiretype{n}  & \setwiretype{n}   & \setwiretype{n}  & \setwiretype{n}  & \setwiretype{n}  &  \setwiretype{n} &  \setwiretype{n} & \setwiretype{n}  & \setwiretype{n} & \setwiretype{n} & \setwiretype{n} & \setwiretype{n} & \setwiretype{n} & \setwiretype{n} & \mycdots\\
\lstick[4]{\textsc{\large $n$}} 
  && \qw & \targ{} 
  & \gate[9,disable auto height]{\textsc{\large $\|Y^{1}\|$}} 
  & \qw & 
  & \gate[9,disable auto height]{\textsc{\large $\|Y^{2\dagger}\|$}}
  & \targ{} 
  & \gate[9,disable auto height]{\textsc{\large $\|Y^{2}\|$}} 
  & \qw  & \qw  & \qw &   \gate[9,disable auto height]{\textsc{\large $\|Y^{k\dagger}\|$}}  & \targ{} & \gate[9,disable auto height]{\textsc{\large $\|Y^{k}\|$}} & &\gate[4,disable auto height]{\textsc{\large QFT}} & \qw & \gate{X} & \ctrl{5}\wire[u][2]{q} &  \control{} \wire[u][2]{q}  & \qw & \gate{X}  & \gate{X} &  \ctrl{6}\wire[u][2]{q} & \control{}\wire[u][2]{q} & \qw & \gate{X} & \qw & \qw & \qw & \qw & \qw & \qw & \qw  & \gate[4,disable auto height]{\textsc{\large IQFT}} &  \meter{}\\
                && \qw  & \qw & \qw & \qw && \qw & \qw & \qw & \qw  & \qw  & \qw & \qw  & \qw & \qw & \qw & \qw & \qw  & \qw & \qw & \qw & \qw & \qw  & \qw & \qw & \qw & \qw & \qw & \qw &  \qw & \qw& \qw & \qw & \qw & \qw & \qw &  \meter{}\\
                &&& \qw & \qw & \qw & \qw & \qw & \qw & \qw & \qw& \qw & \qw &  \qw  & \qw & \qw & \qw & \qw & \qw & \qw & \qw   & \qw & \qw & \qw & \qw & \qw   & \qw & \qw & \qw & \qw & \gate{X}& \ctrl{5}\wire[u][4]{q}  & \control{}\wire[u][4]{q}  & \qw & \gate{X} & \qw & \qw  &  \meter{}\\
                && &\qw & \qw & \qw & \qw & \qw & \qw & \qw & \qw& \qw  & \qw & \qw & \qw  & \qw & \qw & \qw & \qw & \qw & \qw & \qw  & \qw & \qw & \qw & \qw & \qw & \qw & \qw & \qw & \qw & \qw  & \qw & \qw & \qw & \qw & \qw  &  \meter{}\\
                 & \setwiretype{n}& \setwiretype{n} &\setwiretype{n} & \setwiretype{n} & \setwiretype{n} & \setwiretype{n} & \setwiretype{n} & \setwiretype{n} & \setwiretype{n}   & \setwiretype{n} & \setwiretype{n} & \mycdots \setwiretype{n} & \setwiretype{n}  & \setwiretype{n}  & \setwiretype{n}  & \setwiretype{n}  & \setwiretype{n}  &  \setwiretype{n}  &  \setwiretype{n} & \setwiretype{n} & \setwiretype{n} & \setwiretype{n} &  \setwiretype{n} & \setwiretype{n} & \setwiretype{n} & \setwiretype{n} & \setwiretype{n} & \setwiretype{n} & \mycdots\\
\lstick[4]{\textsc{\large $m$}} 
  && \qw & \qw & \qw  & \qw & \qw & \qw & \qw & \qw & \qw & \qw  & \qw & \qw & \qw & \qw && \gate[4,disable auto height]{\textsc{\large QFT}} & \qw & \gate{X} & \control{} & \qw &\control{}\wire[u][7]{q}  & \gate{X}  & \qw & \qw & \qw & \qw & \qw & \qw & \qw & \qw& \qw & \qw & \qw & \qw  &  \gate[4,disable auto height]{\textsc{\large IQFT}} & \meter{}\\
                &&& \qw & \qw & \qw & \qw & \qw & \qw & \qw & \qw & \qw & \qw & \qw& \qw  & \qw & \qw  & \qw & \qw & \qw & \qw  & \qw & \qw & \qw & \gate{X} & \control{}& \qw & \control{}\wire[u][8]{q}  & \gate{X} & \qw & \qw & \qw & \qw & \qw & \qw & \qw & \qw  & \meter{}\\
                && \qw & \qw & \qw & \qw & \qw & \qw & \qw & \qw & \qw & \qw & \qw & \qw& \qw & \qw & \qw & \qw & \qw & \qw & \qw && \qw  & \qw & \qw  & \qw & \qw & \qw & \qw & \qw & \gate{X} & \control{} & \qw & \control{}\wire[u][9]{q} & \gate{X} & \qw & \qw  & \meter{}\\
                &&& \qw & \qw & \qw & \qw & \qw & \qw & \qw & \qw &  \qw & \qw & \qw & \qw & \qw & \qw & \qw & \qw & \qw & \qw & \qw  & \qw & \qw  & \qw & \qw & \qw & \qw & \qw & \qw & \qw & \qw & \qw & \qw & \qw & \qw & \qw  & \meter{}
\end{quantikz}}
\caption{Three-dimensional Sequential-QFL: The diagram comprises three registers containing $k,n$, and $m$ qubits, which facilitate the encoding of a three-dimensional matrix $Y \in \mathbb{R}^{n\times m\times k}$ in the unary basis, here grouped with the red block. The QFT boxes denote the Quantum Fourier Transform applied to the two lower registers to transform both the $i$ and $j $dimensions of $\mathbb{R}^{n\times m}$. The variational quantum circuit (blue block), is applied to appropriately modify a selected pair of Fourier modes, and the Inverse QFT is employed to return to the spatial domain. In the end, we measure probability amplitudes.}

\label{3dimension}
\end{figure*}

\section{Methodology}\label{methodology}
In this work, we introduce our PH-QFNO utilizing PennyLane interface, inspired by the sequential quantum circuit~\cite{jain2023quantumfouriernetworkssolving} to simulate the Burgers' equation, and its 2D generalization applied to the Navier-Stokes equation. The proposed architecture can be extended to multiple dimensions, enabling the future expansion of this approach to a three-dimensional domain.

\subsection{Unary Encoding of Three-dimensional matrices}
To encode a three-dimensional matrix $Y\in\mathbb{R}^{m\times n\times k}$ with entries \(y^l_{ij}\), where the upper index \(l\) denotes the third dimension and the indices \(i,j\) correspond to the two-dimensional matrix at level \(l\), the procedure naturally extends from the two-dimensional case described in Sec.~\ref{unaryencoding}, but one additional register is required, as illustrated in the red block of Fig.\ref{3dimension}.

The first step is to encode the Frobenius norm of $k$ matrices $(m\times n)$ into the first register. Then, the lower two registers are used to encode each 2D matrix $Y^l\in\mathbb{R}^{m\times n}$, following Section~\ref{unaryencoding}. By leveraging the unitarity of the encoding procedure, the final state is obtained as:
\begin{align*}
        \ket{Y} = \frac{1}{\|Y\|}\left(\sum_{i,j,l=1}^{m,n,k}y^l_{ij}\ket{e_i}\ket{e_j}\ket{e_l}\right).
\end{align*}
The mathematical proof is straightforward following the diagram in Fig.~\ref{3dimension}.  

\subsection{Three dimensional Unary Quantum Fourier Tranform}
To Fourier transform the embedded amplitudes, the procedure directly follows from Section~\ref{unaryqft}, and we need to permute both the dimensions $i$ and $j$ of the initial three-dimensional matrix $Y\in\mathbb{R}^{m\times n\times k}$, 
\begin{equation}\label{qftstate3}
    \ket{Y} = \frac{1}{\|Y\|}\left(\sum_{i,j,l=1}^{m,n, k} x_{\lfloor i ,j\rfloor}^l\,\ket{e_i}\ket{e_j}\ket{e_l}\right).
\end{equation}
and apply the $\hat{Q}_{FT}$ operator to the second and third register, mimicking the classical two-dimensional FFT, as shown in Fig.~\ref{3dimension}. 
Hence, after this computation, we get: 
{\small
\begin{equation}
    \begin{gathered}\label{qftstate4}
        \frac{1}{\|Y\|}(\hat{Q}_{FT}\otimes\hat{Q}_{FT}\otimes\mathbb{I})\left(\sum_{i,j,l=1}^{m,n, k} y_{\lfloor i ,j\rfloor}^l\,\ket{e_i}\ket{e_j}\ket{e_l}\right)=\\\frac{1}{\|Y\|}\left(\sum_{i,j,l=1}^{m,n, k} \hat{y}_{ i ,j}^l\,\ket{e_i}\ket{e_j}\ket{e_l}\right).
    \end{gathered}
\end{equation}}
which correctly performs the one dimensional QFT over both the dimensions of the $k$ matrices $Y^l\in \mathbb{R}^{m\times n}$, as in the classical two-dimensional FFT.

\subsection{Variational Quantum Gates}\label{varquantum}
In this subsection, we describe the variational quantum gates designed for both 2D and 3D inputs. The 2D structure is used as a foundation upon which we constructed the new 3D architecture.

After the Fourier transform, the state vector for both the 2D and 3D circuits can be represented as the following superposition in the unary basis:
\begin{align}
    \ket{\hat{X}}=\frac{1}{\|X\|}\displaystyle\sum_{i,j=1}^{m,n}\hat x_{ij}\ket{e_i}\ket{e_j}\quad 2D  \label{eq11}\\
    \ket{\hat{Y}}=\frac{1}{\|Y\|}\displaystyle\sum_{i,j,l=1}^{m,n,k}\hat y^l_{ij}\ket{e_i}\ket{e_j}\ket{e_l}\quad 3D \label{eq12}
\end{align}
To emulate the classical learning matrices $W_K$ and $W_{K_1K_2}$ from FNO \cite{FNO}, we focus on first modes $K=1$ and $(K_1,K_2)=(1,1)$. However, the approach generalizes naturally to any $ K \in [1,n]$ for 2D and any $ K_x,K_y \in [1,m],[1,n]$ for 3D. Fig.~\ref{3dimension} illustrates the 3D quantum learning gates in our hybrid model.

We begin by isolating the elements in Eqs.~\ref{eq11} and~\ref{eq12} corresponding at the first quantum modes, that will be modified by the learning matrices, by applying $X$ gates to the first qubit of second register for 2D, and to the second and third registers for 3D:
{\small
\begin{align}
    &\frac{1}{\|X\|}\displaystyle\sum_{i=1}^{m}\left(\hat x_{i1}\ket{e_i}\ket{e_0}  + \sum_{j=2}^{n}\hat{x}_{ij}\ket{e_i}X_1\ket{e_j} \right) \label{eq13}\\
    &\frac{1}{\|Y\|}\displaystyle\sum_{l=1}^{k}\left(\hat y^l_{11}\ket{e_0}\ket{e_0}  +\sum_{i,j=2}^{m,n}\hat{y}_{ij}^l X_1\ket{e_i} X_1\ket{e_j}\right) \ket{e_l} \label{eq14}
\end{align}
}
where the lower index in $X_1$ stands for the first qubit.

Now, following the \emph{Sequential} circuit described in Ref.~\cite{jain2023quantumfouriernetworkssolving}, we proceed by applying to the first register the learning blocks $C_ZP^\dagger(-\bar{\theta}_1)C_ZP(\bar{\theta}_1)$ for 2D and $CC_ZP^\dagger(-\bar{\theta}_{11})CC_ZP(\bar{\theta}_{11})$ for 3D, where  $C_Z$ and $CC_Z$ are sequences of controlled- and double controlled-Z gates and $P(\bar{\theta}_1)$ is the Quantum Orthogonal Layer described in Ref.~\cite{Orthogonal, Landman2022quantummethods}, with lower indices referring to the first mode. The number of learning parameters depends on the number of the $n$ wires of the first register, in particular $\bar{\theta} \in \mathbb{R}^D$ where $D=\frac{n}{2}\log(n)$.

\noindent \textbf{2D Control-Z gate:} 

After the application of the parametrized learning block, Eq.~\ref{eq13} becomes:
{\small
\begin{align*}
    &\frac{1}{\|X\|}\displaystyle\sum_{i=1}^m \left( \hat{x}_{i1}\left(P^\dagger(-\bar{\theta}_1)P(\bar{\theta}_1)\right) \ket{e_i}\ket{e_0} +\right.\\
    &\left.  +\sum_{j=2}^n\hat{x}_{ij}\left( C_ZP^\dagger(-\bar{\theta}_1)C_ZP(\bar{\theta}_1) \right)\ket{e_i}X_1\ket{e_j}\right),
\end{align*}}
in which the controlled-Z do not affect the $\hat{x}_{i1}$ amplitudes.

By property $C_ZP^\dagger(-\bar{\theta}_1)C_ZP(\bar{\theta}_1)=\mathbb{I}$, applying the $X_1$ gate again on second register, we can obtain:
\begin{equation}
    \frac{1}{\|X\|}\displaystyle\sum_{i=1}^m\left(\hat{\xi}_i(\bar{\theta}_1)\ket{e_i}\ket{e_1} + \sum_{j=2}^n\hat{x}_{ij}\ket{e_i}\ket{e_j}  \right)
\end{equation}
where $\xi_i(\bar{\theta}_1)=\sum_sW_{si}^Q(\bar{\theta}_1)x_{s1}$ and $W^Q_1=W^Q(\bar{\theta}_1)=P^\dagger(-\bar{\theta}_1)P(\bar{\theta}_1)$ which is the quantum implementation corresponding to the matrix $W_1$ used in the FNO \cite{FNO}. To generalize the transformation to a higher number of modes $K$, the operations on the second register are repeated sequentially. The resulting quantum state is given by:
\begin{equation}
    \frac{1}{\|X\|}\displaystyle\sum_{i=1}^m\left(\sum_{j=1}^K\hat{\xi}_i(\bar{\theta}_j)\ket{e_i}\ket{e_j} + \sum_{j=K+1}^n\hat{x}_{ij}\ket{e_i}\ket{e_j}  \right)
\end{equation}
conserving correctly the unary basis.

\noindent \textbf{3D Double Control-Z gate:} 
In the three-dimensional Sequential quantum circuit, we need to deal with all the possible combinations of the $\sum_{i,j=2}^{m,n}\hat{y}_{ij}^l X_1\ket{e_i} X_1\ket{e_j}$ term in Eq.~\ref{eq14}, to preserve the identity
\begin{equation}\label{identity}
    CC_ZP^\dagger(-\bar{\theta}_{ij})CC_ZP(\bar{\theta}_{ij}) =\mathbb{I}
\end{equation}
and to conserve the unary basis.
To achieve this, we introduce the following quantum block: 

\begin{center}
\begin{quantikz}
& \gate{Z} \gategroup[4,steps=3,style={dashed,rounded
corners,fill=blue!10, inner
xsep=1pt},background,label style={label
position=below,anchor=north,yshift=-0.2cm}]{{\sc
$CC_{Z}$}} & \gate{Z} & \gate{Z} & \rstick{First register}  \\
& \setwiretype{n} & \setwiretype{n} & \setwiretype{n} & \setwiretype{n} & \setwiretype{n}\\
\lstick{qubit i} & \ctrl{1}\wire[u][2]{q} & \control{}\wire[u][2]{q} &\qw & \rstick{Second register}\\
\lstick{qubit j}& \control{} & \qw &\control{}\wire[u][3]{q} & \rstick{Third register}
\end{quantikz}
\end{center}
that will substitute the simple controlled-Z gate inside the learning layer in the \emph{Sequential} Quantum circuit, conserving all the properties that we need.

Now, we can apply the block  $CC_ZP^\dagger(-\bar{\theta}_{ij})CC_ZP(\bar{\theta}_{ij}) $ to the first register, as in the 2D case, and get 
\begin{align*}
    &\frac{1}{\|Y\|}\displaystyle\sum_{l=1}^{k}\left(\hat y^l_{11}\ket{e_0}\ket{e_0} \left(P^\dagger(-\bar{\theta}_{11})P(\bar{\theta}_{11})\right) \right.\\ &+\left. \sum_{i,j=2}^{m,n}\hat{y}_{ij}^l \left(CC_ZP^\dagger(-\bar{\theta}_{ij})CC_ZP(\bar{\theta}_{ij})\right) X_1\ket{e_i} X_1\ket{e_j}\right) \ket{e_l}.
\end{align*}

Then, using the identity relation in Eq.~\ref{identity} and applying the $X$ gates to restore the initial states, we have 
\begin{align*}
\frac{1}{\|Y\|}\displaystyle\sum_{l=1}^{k}\left(\hat{\chi}^l(\bar{\theta}_{11})\ket{e_0}\ket{e_0} + \sum_{i,j=2}^{m,n}\hat{y}_{ij}^l  \ket{e_i} \ket{e_j}\right) \ket{e_l}.
\end{align*}
where, as before, $\chi^l(\bar{\theta}_{11})=\sum_{s=1}^kW_{s}^{Ql}(\bar{\theta}_{11})\hat y^s_{11}$ and $W^Q_{11}=W^{Q}(\bar{\theta}_{11})=P^\dagger(-\bar{\theta}_{11})P(\bar{\theta}_{11})$ as before.\\
Hence, we can iterate the sequential controlled learning gate, as in Fig.~\ref{3dimension} to cover different number of modes $(K_x,K_y)$ and obtain
\begin{align*}
\frac{1}{\|Y\|}\displaystyle\sum_{l=1}^{k}\left(\sum_{i,j=1}^{K_x,K_y}\hat{\chi}^l(\bar{\theta}_{ij})\ket{e_i} \ket{e_j} + \sum_{\substack{i=K_x+1 \\ j=K_y+1}}^{m,n}\hat{y}_{ij}^l  \ket{e_i} \ket{e_j}\right) \ket{e_l}.
\end{align*}
Finally, we can apply the inverse QFT, IQFT, on the lower registers on both the 2D and 3D quantum circuit to return to the physical domain and measure probability amplitudes to conserve the \texttt{pyTorch} gradient and introduce a non-linearity in the model.

\begin{figure*}[htbp]
    \centering
    \begin{subfigure}[b]{\linewidth}
        \centering
        \includegraphics[width=0.8\linewidth]{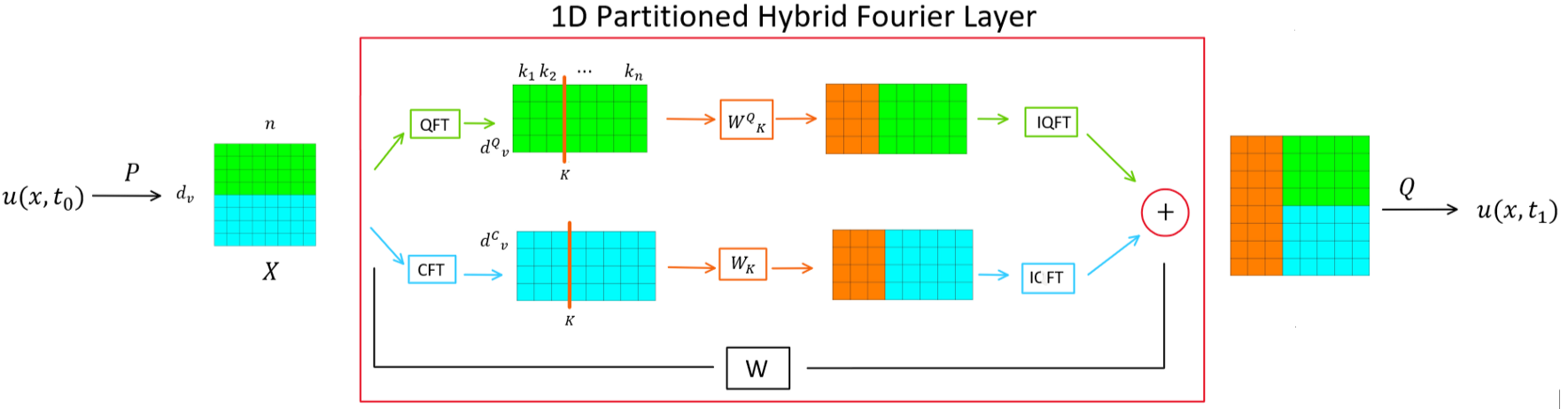}
        \caption{One-dimensional PH-QFNO. The input, \(u(x,t_0) \in \mathbb{R}^n\), is first mapped via a linear transformation \(P(u(x,t_0)) = X \in \mathbb{R}^{d_v \times n}\) which is then partitioned into two submatrices, \(X^Q \in \mathbb{R}^{d_v^Q \times n}\) for the quantum branch and \(X^C \in \mathbb{R}^{d_v^C \times n}\) for the classical branch. Subsequently, the Fourier Layer is split accordingly, where a Fourier transform is initially applied, followed by parameterized learning layers \(W^Q_K\) in the quantum branch and \(W_K\) in the classical branch (with \(K\) denoting the number of modes undergoing parametric multiplication, here colored in orange). $W$, in the black box, is a 2D convolutional layer coming from the FNO. Finally, an inverse Fourier transform restores the solution to the physical domain, and the outputs from both branches are concatenated and further processed by a neural network to recover the original dimension, ultimately yielding the velocity field at a later time \(u(x,t_1)\).}
        \label{FNO1}
    \end{subfigure}
    \hfill
    \vspace{0.1 cm}
    
    \begin{subfigure}[b]{\linewidth}
        \centering
        \includegraphics[width=0.8\linewidth]{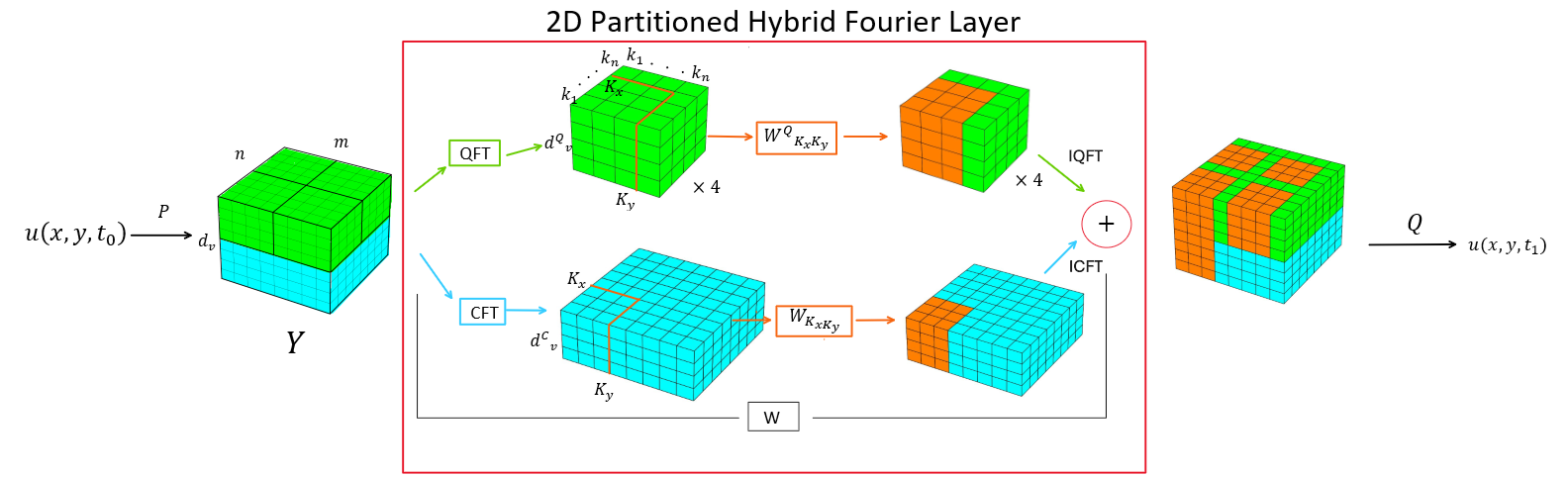}
        \caption{Two-dimensional PH-QFNO. After a linear transformation, the initial velocity field \( u((x,y),t) \in \mathbb{R}^{m \times n} \) is embedded into a higher-dimensional space by introducing an additional dimension \( d_v \). The resulting matrix $Y$ is partitioned into two submatrices, \( Y^Q \in \mathbb{R}^{m \times n \times d_v^Q} \) for the quantum branch and \( Y^C \in \mathbb{R}^{m \times n \times d_v^C} \) for the classical branch. \( Y^Q \) is further divided into four submatrices \( Y_{ij}^Q \), each processed independently by the QFL, analogous to the processing performed on \( Y^C \) by its classical counterpart. The learning gates, denoted by \( W_{K_xK_y}^Q \) in the quantum branch and \( W_{K_xK_y} \) in the classical branch, perform the two-dimensional parametric multiplications on the selected modes (highlighted in orange), while $W$ in the black box, is a 3D convolutional layer. Finally, the inverse Fourier transform is applied and the outputs of the submatrices are concatenated and fed into a neural network that restores the original dimensionality, ultimately yielding the velocity field at a later time.}
        \label{FNO2}
    \end{subfigure}
    
    \caption{Overview of the PH-QFNO architectures: (a) One-dimensional and (b) Two-dimensional PH-QFNO.}
    \label{FNOcombined}
\end{figure*}

\subsection{Partitioned Hybrid Quantum Fourier Neural Operator}
The main challenge in developing the QFNO on a quantum state-vector simulator, such as the one provided by Pennylane, lies in the limited number of qubits that can be efficiently simulated. These simulators must represent the full Hilbert space, making subspace restriction techniques, such as those in Ref.~\cite{jain2023quantumfouriernetworkssolving}, unfeasible. Unary amplitude embedding requires a number of qubits equal to the total input dimension (see Eq.~\ref{unaryencoding}), so we cap the qubit count at 12. To accommodate this constraint, we adapt the computational domain by uniformly sampling \(D_j\) at $8$ points for Burgers' equation and using an \(8 \times 8\) grid for the Navier–Stokes equation.
To address the qubit limitation, we implement a new variant of the QFNO that we called Partitioned Hybrid Quantum Fourier Neural Operator.  

\subsubsection{One-dimensional PH-QFNO }
The entire process is described in Fig.~\ref{FNO1}. The algorithm takes as input a vector \( u_0(x) \) with \( x \in \mathbb{R}^n \), then, after a linear transformation $P$ it is mapped to \(X \in \mathbb{R}^{d_v \times 8}\). The quantum circuit is capable of processing the whole matrix when \(d_v \leq 4\). If \(d_v > 4\), the input $X$ is partitioned into two submatrices:  $X^Q \in \mathbb{R}^{d_v^Q \times 8} \quad \text{with } d_v^Q = 4,$
which is processed by the QFNO, and the remaining part $X_{cl} \in \mathbb{R}^{d_v^C \times 8}$, which is processed by the FNO. The $W$ in Fig.~\ref{FNO1}  represents a global 2D convolutional layer, as part of the classical Fourier Layer. The outputs from these two processing paths are then concatenated. In this framework, the parameter \(d_v\) effectively controls the proportion of the quantum circuit relative to the classical one.

\subsubsection{Two-dimensional PH-QFNO }\label{sub}
This process is depicted in Fig.~\ref{FNO1}. The algorithm takes as input the velocity $u_0(x,y)$ with \( (x,y) \in \mathbb{R}^{m\times n} \) which is then mapped linearly to $Y \in \mathbb{R}^{8 \times 8 \times d_v}.$
As in the previous case, if \(d_v > 4\) the matrix is partitioned into two submatrices, namely $Y^Q \in \mathbb{R}^{8 \times 8 \times d_v^Q  } \quad \text{(with } d_v^Q = 4\text{)},$ and  $Y^{cl} \in \mathbb{R}^{ 8 \times 8 \times d_v^C}.$
However, in this scenario the quantum circuit can process only matrices in \(\mathbb{R}^{4 \times 4 \times 4}\). Therefore, the matrix \(Y_q\) is further subdivided into four submatrices, denoted \(Y_{ij}^Q\), each of which is processed independently with different parameters. This aspect introduces a higher expressiveness in the model with respect to the CFNO.

This slicing technique can be straightforwardly extended to larger domains by partitioning the Fourier-layer input into independently analyzable submatrices. Nevertheless, partitioning the global Fourier operator into \(4\times4\times d^Q_v\) subblocks inevitably induces spatial under‐sampling on an \(8\times8\times d_v\) tensor, effectively truncating the lowest‐frequency modes that a classical FNO would retain.  In PH-QFNO, however, we address this spectral gap through two synergistic mechanisms.  Upon reassembling the four quantum‐processed subblocks, we apply a global classical convolutional layer \(W\), which operates across the full spatial domain to reintroduce long‐range correlations and accelerate convergence.  Moreover, by deploying four independent quantum Fourier operators rather than just a single  one, the model’s total parameter count increases fourfold, endowing the subsequent network layers with sufficient expressive capacity to \emph{fill in} the omitted low frequencies via learned weights.  Empirically on Burgers’ and Navier–Stokes benchmarks, this combination of global classical integration and controlled over-parameterization compensates for the truncated spectrum, delivering accuracy and stability on par with, or in some cases exceeding, that of the full-domain FNO.  

\subsection{Data Parallelism}
To address the computational limitations of state-vector quantum computer simulators, we adot an MPI-based data parallelism strategy~\cite{gropp1999using}. The complete dataset is partitioned and distributed evenly across multiple processes. Each process performs forward and backward propagation independently on its data subset. Before updating the model parameters using the optimizer, gradient synchronization is performed by MPI reduction across all processes to ensure consistent parameter updates. Finally, the loss values from each process are averaged to compute the global loss.


\section{Experimental Setup}
\begin{figure*}[htbp]
    \centering
    \begin{subfigure}[t]{0.35\textwidth}
        \centering
        \includegraphics[width=\linewidth]{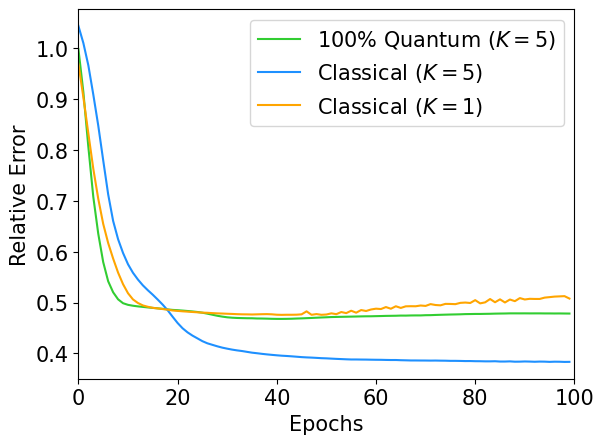}
        \caption{1D Burgers' Equation; relative error comparison between fully quantum and classical models across epochs.}
        \label{fig:relativeerrorplot1d}
    \end{subfigure}%
    \hspace{0.05\linewidth}
    \begin{subfigure}[t]{0.35\textwidth}
        \centering
        \includegraphics[width=\linewidth]{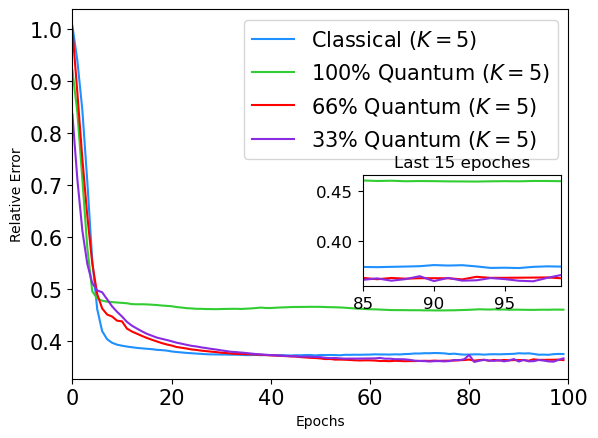}
        \caption{1D Burgers' Equation; relative error comparison between hybrid and classical models.}
        \label{fig:relativeErrorComparisonRatio1d}
    \end{subfigure}

    \vspace{0.2cm}

    \begin{subfigure}[t]{0.35\textwidth}
        \centering
        \includegraphics[width=\linewidth]{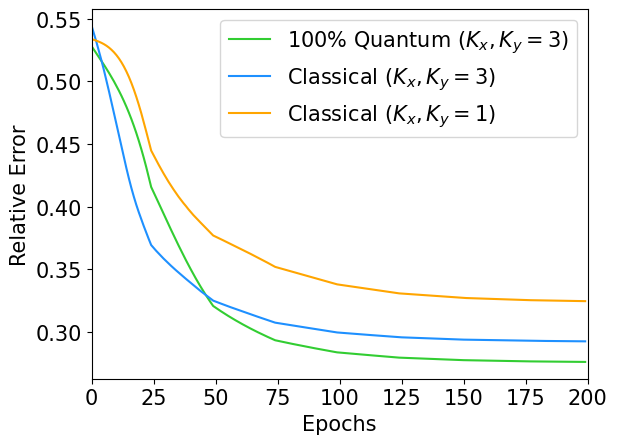}
        \caption{2D Navier–Stokes Incompressible Equations: relative error comparison between fully quantum and classical models.}
        \label{fig:relativeerrorplot2d}
    \end{subfigure}%
    \hspace{0.05\linewidth}
    \begin{subfigure}[t]{0.35\textwidth}
        \centering
        \includegraphics[width=\linewidth]{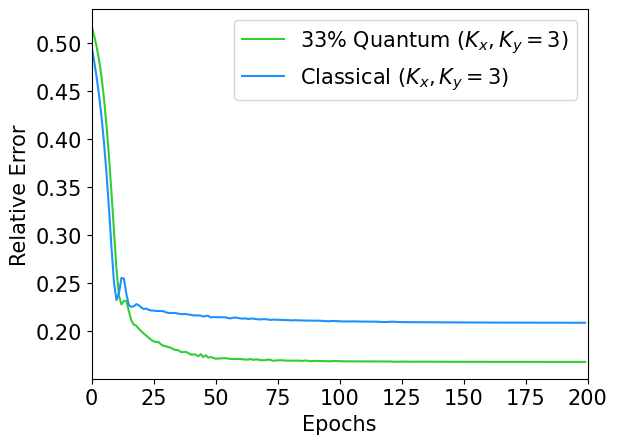}
        \caption{2D Navier–Stokes Incompressible Equation: relative error comparison between hybrid and classical models.}
        \label{fig:relativeErrorComparisonRatio2d}
    \end{subfigure}

    \caption{Relative error comparisons on 1D and 2D PDEs across different model types. In (a) and (c), the dimension of the latent representation of the input function $u(x)$ is set to $d_v = 4$, while in (b) and (d), it is set to $d_v = 12$.}
    \label{fig:relativeerrorcomparisonall}
\end{figure*}
All experiments are conducted on a laptop running Ubuntu 22.04.5 LTS, equipped with an AMD Ryzen 7 5800H CPU and an NVIDIA GeForce RTX 3060 Mobile GPU. Quantum circuits are implemented using PennyLane’s \texttt{default.qubit} state-vector simulator~\cite{bergholm2018pennylane, asadi2024hybrid}, which supports parameterized quantum circuits and integrates with PyTorch via \texttt{qml.qnn.TorchLayer}. Data parallelism is employed by distributing the dataset across five processes in all experiments using \texttt{mpi4py}~\cite{dalcin2021mpi4py}.

We test and verify the PH-QFNO against two test problems:

\textbf{1. Burgers' Equation.} The one-dimensional Burgers' equation is a nonlinear partial differential equation that describes the motion of a viscous fluid, and is formulated as:
{\small
\begin{align*}
    \partial_{t} u(x, t) + \partial_{x}\left(\frac{u^2(x, t)}{2}\right) &= \nu \partial_{xx}u(x,t) \quad x \in (0,1),t \in (0,1] \\
    u(x,0) &= u_{0}(x) \quad x \in (0,1)
\end{align*}}
with periodic boundary conditions where $\nu \in \mathbb{R}_+$ is the viscosity coefficient of the fluid and $u_0\in\mathcal{L}_{\text{per}}^2((0,1);\mathbb{R})$ is the initial condition. The model is expected to learn the operator mapping the initial condition to the solution at t = 1, i.e. $\mathcal{G}^\dagger\,:\,\mathcal{L}_{\text{per}}^2((0,1);\mathbb{R})\to\mathcal{H}_{\text{per}}^s((0,1);\mathbb{R})$ defined by $u_0\to u(\cdot,1)$ for any $s>0$.

\textbf{2. Navier-Stokes Equations.} We consider the vorticity-streamfunction formulation of the two-dimensional Navier-Stokes equations for a viscous, incompressible fluid:
{\small
\begin{align*}
    \partial_{t} w(x, t) + u(x, t)\cdot \nabla w(x,t)&= \nu \Delta w(x,t) + f(x)\\ 
    x \in (0,1)^2, t& \in (0,T]\\
    \nabla \cdot u(x,t)= 0 \quad x &\in (0,1)^2, t \in [0,T]\\
    w(x,0) = w_0(x)\quad x &\in (0,1)^2
\end{align*}}
equipped with periodic boundary conditions where $u\,:\,(0,1)^2\times \mathbb{R}_+ \to\mathbb{R}^2$ is the velocity field, $w$ is the out-of-plane component $(0,0,w)$ of the vorticity field $\nabla\times u\,:\,(0,1)^2\times \mathbb{R}\to\mathbb{R}^3$, $f$ is the forcing factor and $\nu$ the viscosity which has been set equal to $1e-3$ and we fix the resolution to be $8\times 8$ for both training and testing. We aim to learn the solution operator $\mathcal{G}^\dagger\,:\,\mathcal{L}^2_{\text{per}}((0,1)^2;\mathbb{R})\to\mathcal{H}^s_{\text{per}}((0,1)^2;\mathbb{R})$ defined by $w_0\to w(\cdot,t)$ with $t_0 =30$ and $t=31$ region in which the dynamic of the fluid becomes more chaotic to strengthen the validity of our approach. The initial vorticity field \(w(x,0)\) is sampled from the Gaussian measure $\mu = \mathcal{N}\bigl(0,\;7^{3/2}(-\Delta + 49I)^{-2.5}\bigr)$ with periodic boundary conditions; for further details see Ref.~\cite{FNO}.

In addition to incompressible flow, we also evaluate the HP-QFNO on solving the compressible Navier-Stokes equations for the Kelvin-Helmholtz configurations. The dataset is generated by solving the compressible Euler equations and introducing small random perturbations into the initial conditions for density, velocity, and pressure, thereby producing a time-series dataset, capturing the evolution of the flow dynamics. This dataset, which encompasses a range of scenarios for the initial conditions and subsequent flow evolution, is subsequently employed to evaluate the performance of our QFNO in approximating the operator mapping the density at time $t=0$ up to time $t=4$, i.e., $G^\dagger : \mathcal{L}_{\text{per}}^2((0,1);\mathbb{R}) \to C((0,4]; H_{\text{per}}^r((0,1);\mathbb{R}))$.

\subsection{Hybrid Quantum Neural Network Setup}
In this subsection, we describe the quantum neural network setup used in our experiments. For every simulation, we employ 12 qubits, which is the maximum allowed by PennyLane to perform simulations within a reasonable time frame.
Training of the proposed architecture was performed using backpropagation and the Adam optimizer.
\begin{table}[htbp]
  \centering
  \resizebox{\columnwidth}{!}{%
    \begin{tabular}{cccccccc}
      \toprule
      & & \multicolumn{2}{c}{\textbf{Burger}} & \multicolumn{2}{c}{\textbf{Navier Stokes}} & \multicolumn{2}{c}{\textbf{Kelvin Helmholtz}}\\
      \cmidrule(lr){3-4} \cmidrule(lr){5-6} \cmidrule(lr){7-8}
      qubits & Hybridization  & $d_v$ & $d_v^Q$  & $d_v$ & $d_v^Q$  & $d_v$ & $d_v^Q$ \\
      \midrule
      12 & 100\%       &  4  & 4   &  4  & 4  &  4  & 4 \\
      12 & 100\%       &  12  & $4\times3$  &  /  & / &  /  & / \\
      12 & 66\%       & 12   & $4\times2$   &  /  & / &  /  & / \\
      12 & 50\%       & /   & /   &  /  & / &  8  & 4 \\
      12 & 33\%       & 12   & 4  &  12  & 4 &  /  & / \\
      \bottomrule
    \end{tabular}}
  
  \caption{Configurations for the latent dimensionality \(d_v\) and \(d_v^Q\) for various PDE problems. For Burger equation, $4\times3$ and $4\times2$ indicates that the quantum circuit partitions the input matrix of size $12\times8$ respectively into three or two submatrices of size $4\times8$, processed sequentially by the QFNO, while the others, if present, by the FNO.}
  
  \label{dimensionality}
\end{table}

\noindent \textbf{1. One-Dimensional Burgers' Equation:}
The network input is constructed by concatenating the initial velocity field \(u(x,0)\) with the spatial coordinate \(x\), where \(x \in \mathbb{R}^8\). The first layer of the network maps this two-dimensional input into a higher-dimensional latent space via \texttt{torch.nn.Linear(2, \(d_v\))}, where the parameter \(d_v\) will determine the hybrid proportion of the model. In our implementation, the quantum circuit with 12 qubits can embed a matrix of size \(4 \times 8\). Therefore, if \(d_v\) exceeds 4, a hybrid architecture is required that partitions the matrix into submatrices, as depicted in Fig.~\ref{FNO1}. Table~\ref{dimensionality} summarizes the different hybrid configurations.

After encoding and QFT, the learning gates are applied to the first \(K\) modes as described in Section~\ref{varquantum}. Finally, the output stage of the network is a single-layer linear neural network that maps the latent representation to the output velocity field \(u(x,t=1)\).

Regarding the number of learning parameters, for the classical Fourier layer the parameter count is $K \cdot (d_v^C)^2$, where $d_v^C = d_v - d_v^Q,$ and for the Quantum Fourier layer the count is $K \cdot \frac{d_v^Q}{2}\log d_v^Q$.
At equal number of modes \(K\), the classical Fourier Neural Operator exhibits a quadratic dependence on \(d_v\), which renders the classical circuit more expressive. In contrast, the Quantum Fourier layer employs quantum orthogonal layers with fewer parameters, resulting in a loss of expressiveness but yielding gains in training stability and convergence speed, as observed in our experiments.

\noindent \textbf{2. Two-Dimensional Navier Stokes Equations:}
The input is constructed by concatenating \([u(x,y),\, x,\, y]\), with \(x,y \in \mathbb{R}^8\). The first operation is implemented as \texttt{torch.nn.Linear(3, \(d_v\))}, where, as in the 1D case, \(d_v\) regulates the hybrid proportion of the model. We consider the configurations described in Table \ref{dimensionality}.

For the two-dimensional problems, the number of learning parameters in the classical Fourier layer is given by $K_1 \cdot K_2 \cdot (d_v^C)^2,$ and for the Quantum Fourier layer the number is $K_1 \cdot K_2 \cdot \frac{d_v^Q}{2}\log d_v^Q,$ which brings to the same conclusions of before.

\subsection{Evaluation Metrics}
Model performance is evaluated using the relative error on test sets, as defined in \eqref{relativeerror}, following standard practice in related work. Consistent with prior studies on the classical FNO~\cite{FNO}, we use the same datasets with training/testing splits of 500/50 for Burgers’ equation and 100/25 for the Navier–Stokes equation. For the Kelvin–Helmholtz problem, we generate a new dataset with a 20/5 split. To address qubit limitations in \texttt{PennyLane}, the spatial domain is downsampled to $8$ points for Burgers’ equation and an \(8 \times 8\) grid for the Navier–Stokes equation.

\begin{equation}
    \text{RelError}(\mathbf{y_{pred}}, \mathbf{y_{true}}) = \frac{\|\mathbf{y_{pred}}-\mathbf{y_{true}}\|_2}{\|\mathbf{y_{true}}\|_2}
    \label{relativeerror}
\end{equation}

Experiments include comparative analysis between PH-QFNO  and FNO under varying hybridization rates and multiple configurations of the modal parameters \( K \) and \( (K_x, K_y) \) within the learning layers \( W^Q \) and \( W^K \). Additionally, we conduct a comparative analysis between the quantum circuit implementation and its original classical counterpart by evaluating their robustness to input noise. The similarity we adopted, aka cosine similarity, is quantified using the following metric:

\begin{equation} 
\text{Similarity}(\mathbf{\tilde{x}}, \mathbf{x}) = \frac{1}{B} \sum_{b=1}^{B} \frac{\mathbf{\tilde{x}}_b \cdot \mathbf{x}_b}{\|\mathbf{\tilde{x}}_b\|_2 \cdot \|\mathbf{x}_b\|_2} \label{similarityeq} 
\end{equation}
where $\mathbf{x}_b$ denotes the output corresponding to the $b$-th clean input sample, and $\mathbf{\tilde{x}}_i$ denotes the output generated from the same input with added noise. The variable $B$ represents the batch size, which is set to 5 in our experiments. This similarity metric allows us to quantitatively assess the effect of noise on the output and evaluate the comparative resilience of the quantum and classical models.



\section{Results}\label{results}
This section analyses our proposed quantum algorithms for solving PDEs, see Table~\ref{dimensionality} to have more details about each hybrid model.

\begin{figure}[htbp]
    \centering
    \includegraphics[width=0.8\linewidth]{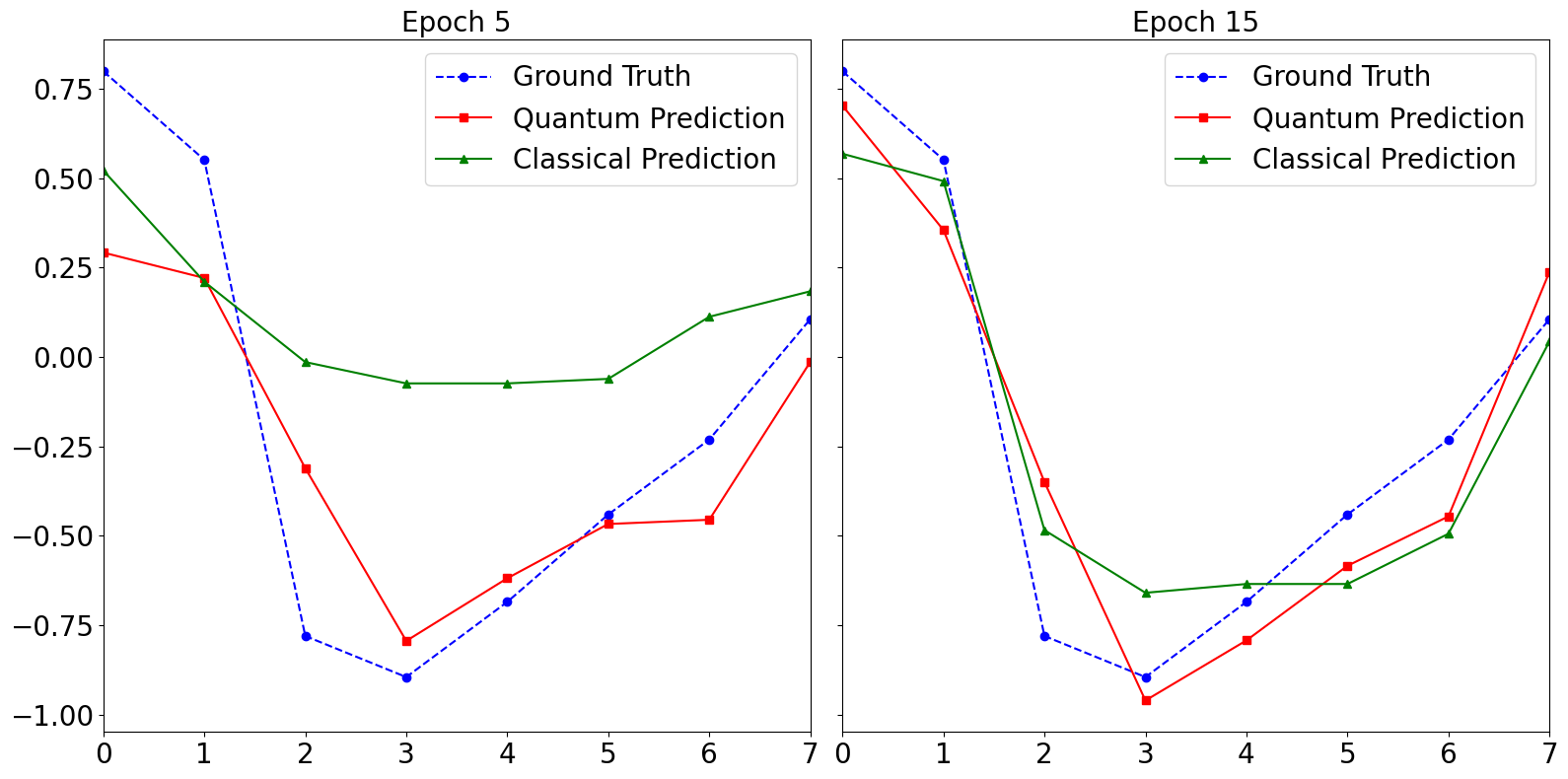}
    \caption{Predictions of the $u$ profile obtained by FNO and PH-QFNO at two different epochs during the training. The parameter $K$ is set to 5 for both the models.}
    \label{fig:prediction100QuantumClassical5modes}
\end{figure}

First, we show the results relative to the solution of the Burger's equation. Fig.~\ref{fig:relativeerrorplot1d} shows the evolution of the relative error on the test dataset during the training of different neural operators. Specifically, we compare the FNO with $K=1  \text{ and } 5$ and our PH-QFNO with $100\%$ quantum hybridization. As illustrated in Fig.~\ref{fig:prediction100QuantumClassical5modes}, where the solution is depicted at two early epochs, the fully quantum model exhibits a significantly faster convergence, rapidly adapting to the ground truth curve relative to the classical FNO models. However, it achieves relative errors comparable to the one-mode classical model by the end of the training. Furthermore, it avoids overfitting with respect to the 1-mode classical model, highlighting that the use of Orthogonal Quantum Layers~\cite{Orthogonal} entails a trade-off between a potential loss in expressiveness and improved training stability. Nevertheless, the five-modes classical model achieves superior performance compared to the others, a result that can be attributed to its higher number of trainable parameters. 

As a second experiment, we study how the degree of quantum layers in the PH-QFNO impacts the accuracy of the results. Fig.~\ref{fig:relativeErrorComparisonRatio1d} shows the usage of classical computation together with quantum counterpart largely enhances the accuracy of the PH-QFNO for the Burger's equation test case: the presence of classical computation (red and purple lines) leads to a smaller relative error compared to classical FNO. 

\begin{figure}[ht]
\centering
    \begin{subfigure}[b]{0.8\linewidth}
        \centering
        \includegraphics[width=\textwidth]{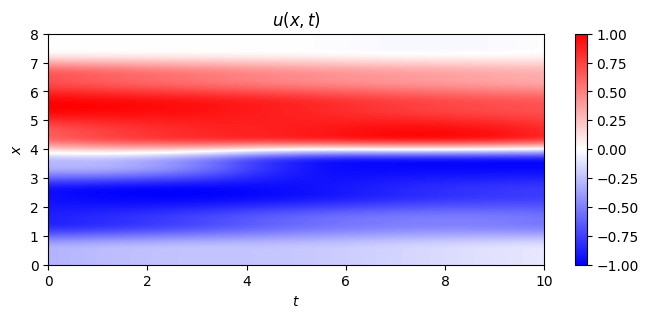}       
    \end{subfigure}
    \begin{subfigure}[b]{0.8\linewidth}
        \centering
        \includegraphics[width=\textwidth]{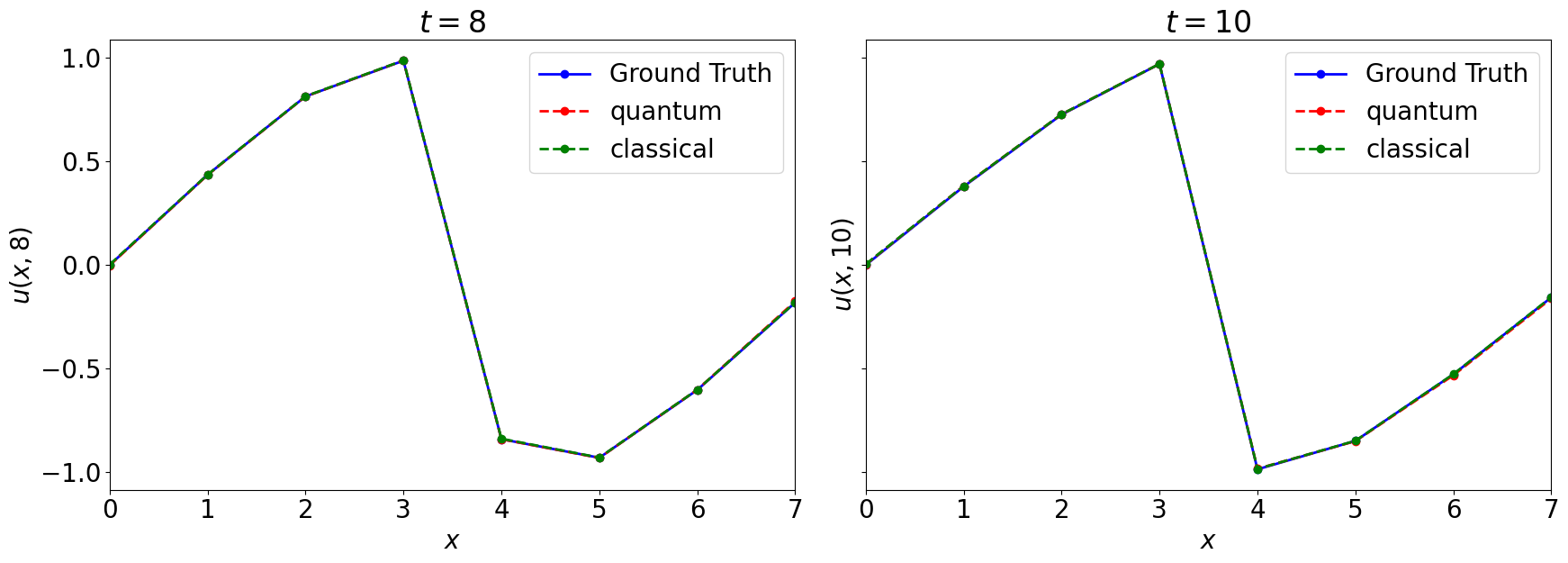}
              
    \end{subfigure}
    \caption{\textbf{Top panel:} Contourplot of the $u$ evolution for 1D Burger with a sinusoidal initial condition. 
    \textbf{Bottom panel:} Comparison between ground truth (blue line), classical prediction (dashed green line) and fully quantum model (dashed red line) over the last two times steps for 1D Burger with sinusoidal initial condition.}
    \label{evolution}
\end{figure}

Furthermore, a third experiment has been conducted. As shown in Fig. \ref{evolution}, an additional analysis was performed to reproduce the Burgers’ shock with $u(x,0)=-\sin(\pi x)$ as initial condition, following the approach presented in \cite{pinns}. We performed the training on this trajectory and reproduced the operator mapping the velocity up to time $t=4$ to the velocity at later times ($T>4$), i.e., $G^\dagger : C([0,4]; H_{\text{per}}^r((0,1);\mathbb{R})) \to C((4,10]; H_{\text{per}}^r((0,1);\mathbb{R}))$.
Both the fully quantum and classical model demonstrated a good adaptation to the shock, yielding a relative error of $10^{-4}$.

\begin{figure}[htbp]
    \centering
    \begin{subfigure}[t]{\linewidth}
        \centering
        \includegraphics[width=\textwidth]{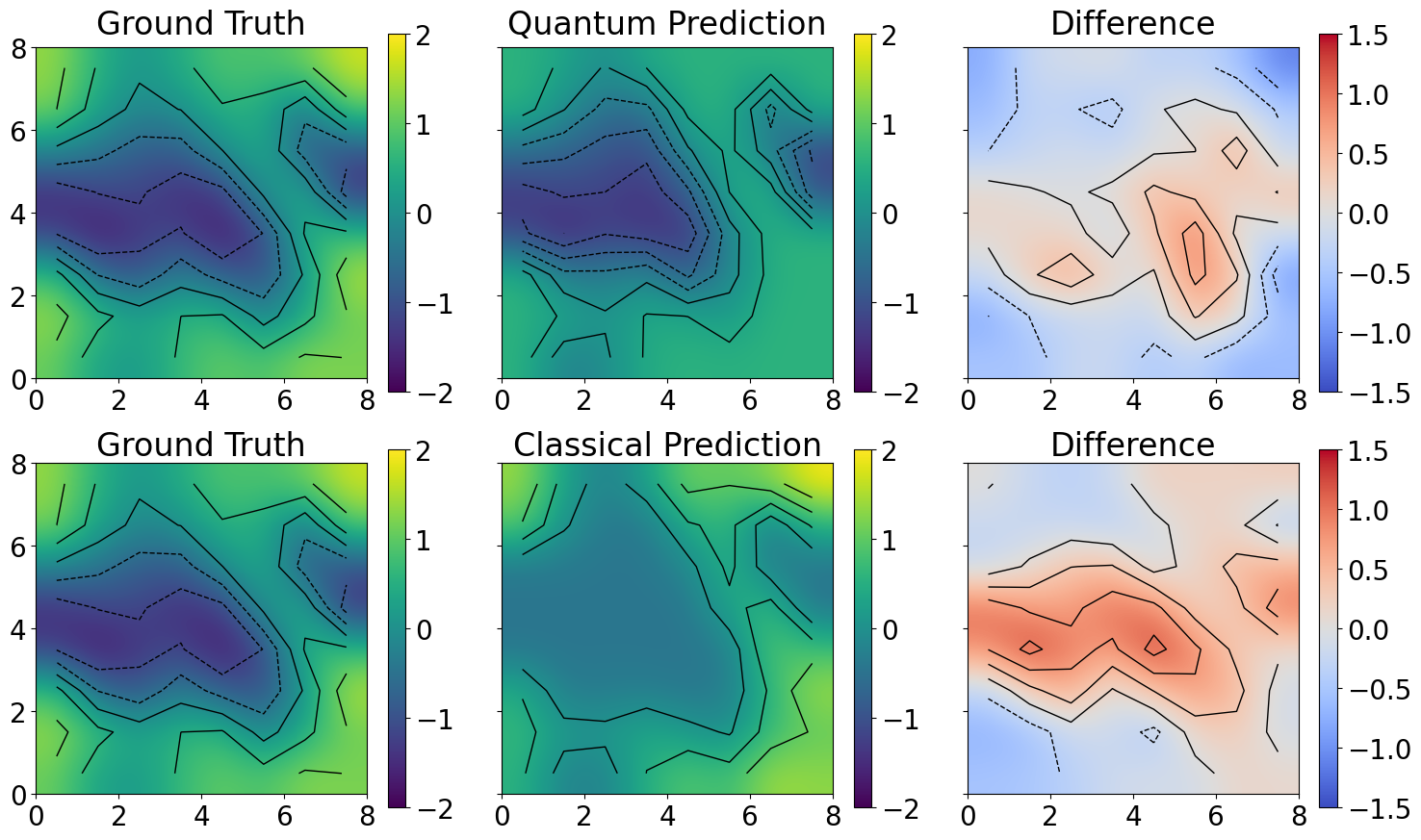}
        \caption{\textbf{Top Panel:} Fully quantum. \textbf{Bottom Panel:} classical FNO. Latent dimensionality $d_v=4$.}
    \end{subfigure}

    \hfill
    \vspace{0.25 cm}
    
    \begin{subfigure}[t]{\linewidth}
        \centering
        \includegraphics[width=\textwidth]{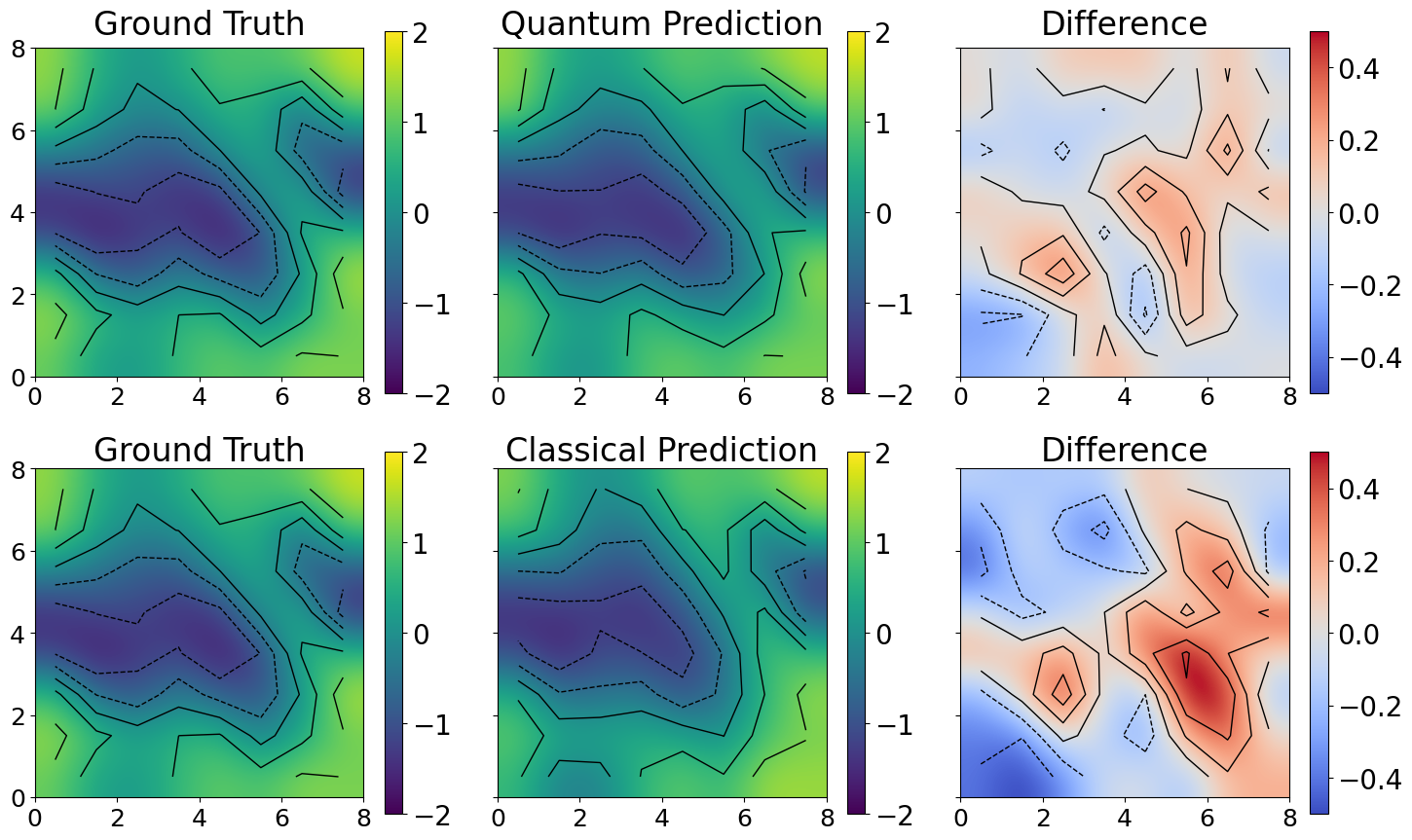}
        \caption{\textbf{Top Panel:} 33\% hybrid quantum. \textbf{Bottom Panel: } classical FNO. Latent dimensionality $d_v=12$.}
    \end{subfigure}
    \caption{Vorticity field of the solution for the incompressible Navier-Stokes problem with viscosity $\nu=1e-3$. Ground truth on the left column, predictions in central position and their difference on the right.}
    \label{ns-errorpred}
\end{figure}

Then, we present the results relative to the two-dimensional Incompressible Navier Stokes equation. In Fig.~\ref{fig:relativeerrorplot2d} we see that the quantum model outperforms both classical FNO, reaching a lower relative error by the end of training. This may be attributed to the fact that the two-dimensional quantum model is designed with a higher number of parameters than its classical counterpart, as explained in Sec. \ref{sub}. Then, Fig.~\ref{fig:relativeErrorComparisonRatio2d} shows that the 33\% PH-QFNO, outperforms the FNO, exhibiting behavior similar to that observed in the one-dimensional case. Indeed, to have a more clear understanding of the results, we provide in Fig. ~\ref{ns-errorpred} a graphical representation of the prediction at time \( t = 1 \) over the two-dimensional domain. It can be observed that the purely quantum model yields more accurate predictions than its classical counterpart. However, the best performance is achieved by the hybrid architecture, which benefits from a larger number of trainable parameters, thereby attaining higher accuracy.  

\begin{figure*}[htbp]
    \centering
    \begin{subfigure}[b]{0.35\textwidth}
        \centering
        \includegraphics[width=\textwidth]{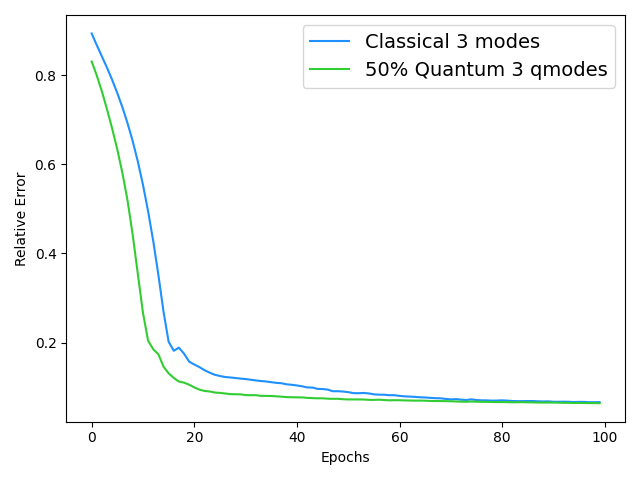}
    \end{subfigure}
    \hspace{0.05\linewidth}
    \begin{subfigure}[b]{0.5\textwidth}
        \centering
        \includegraphics[width=\textwidth]{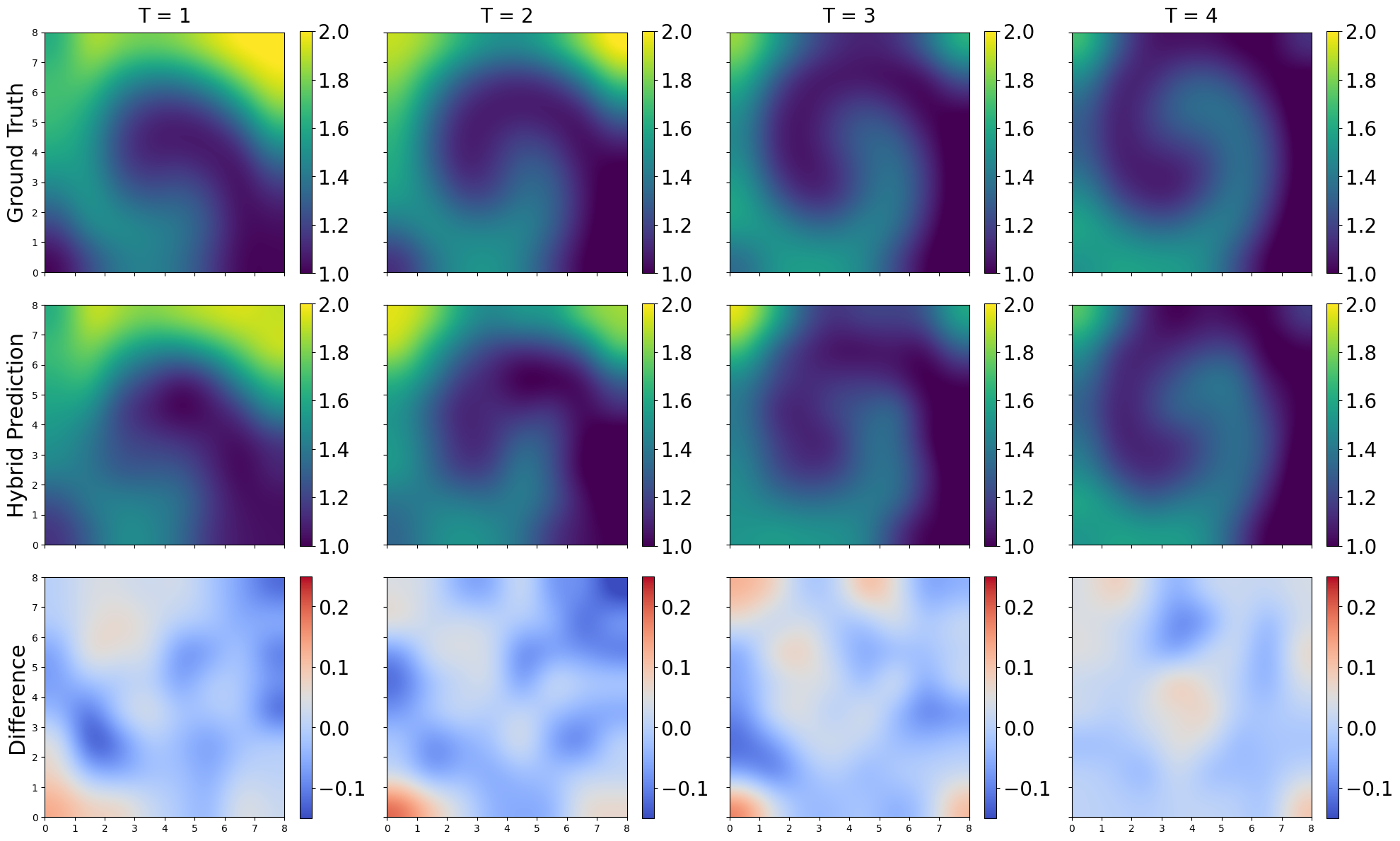}
    \end{subfigure}
    \caption{\textbf{Left Panel:} Relative error variation over training epochs, with the classical model shown in light blue and the 50\% hybrid quantum model in green, both with $K=3$. \textbf{Right Panel:} Evolution of the density field for the two-dimensional Kelvin--Helmholtz equation across multiple time steps: the top row represents the ground truth, the middle row the hybrid quantum prediction, and the bottom row the corresponding relative error.}
    \label{fig:2dpredictions_combined}
\end{figure*}

Additionally, we conducted a further experiment on the compressible Navier Stokes equation, modeling the Kelvin-Helmotz instability.
From Fig.~\ref{fig:2dpredictions_combined} (left), we observe that the 50\% hybrid model converges more rapidly, confirming previous experimental findings, while ultimately achieving an accuracy comparable to that of the classical model. Both the models give accurate predictions, in fact, the relative error across different time steps is very low, as demonstrated in the right-hand panel.

\begin{figure}[htbp]
    \centering
    \includegraphics[width=0.8\linewidth]{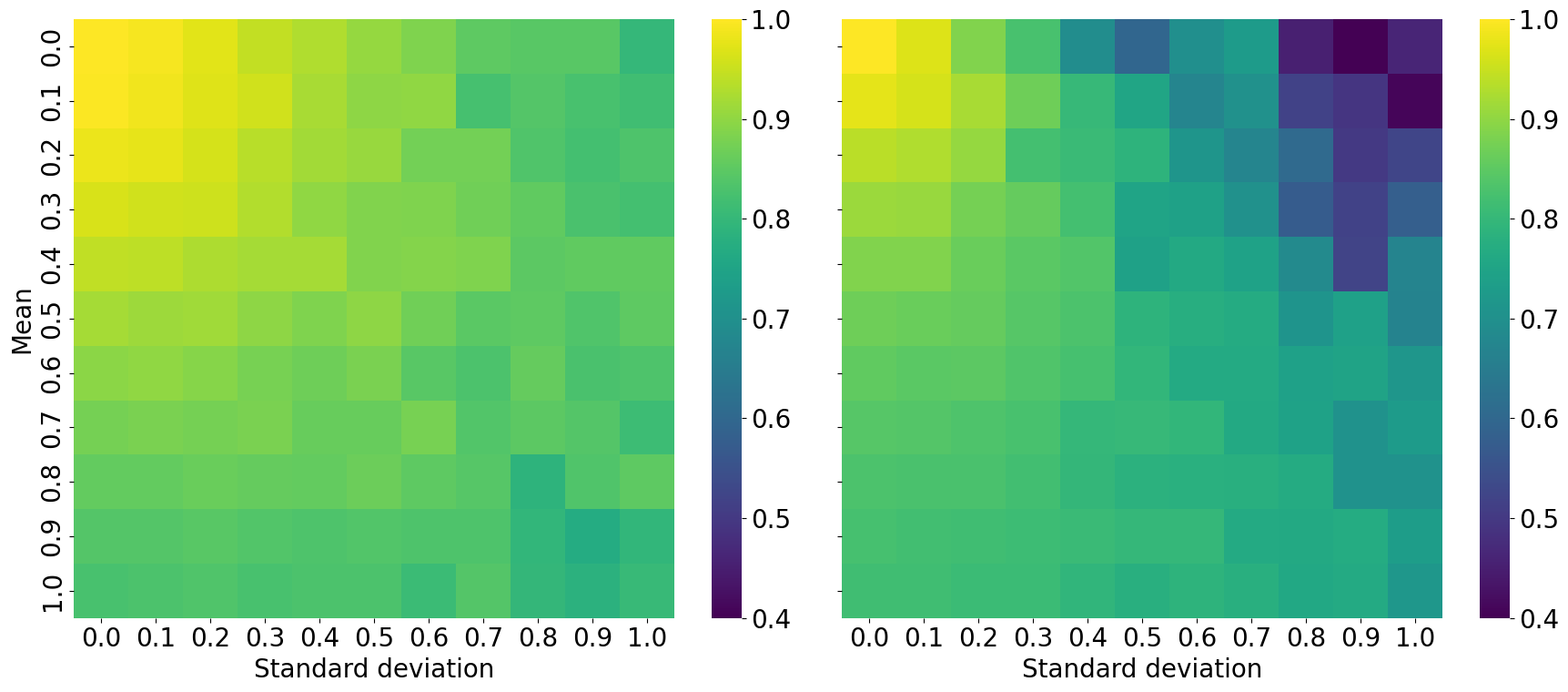}
    \caption{Heatmaps showing the similarity as a function of input noise parameters (mean and standard deviation) for (left) quantum and (right) classical Fourier layer. Both components are evaluated under the same input noise conditions. Brighter colors indicate higher similarity.}
    \label{fig:similarityOnNoise}
\end{figure}

To evaluate the adaptiveness of our quantum Fourier layer, comprising the encoding, QFT, orthogonal layer, and IQFT, we introduce Gaussian noise to the input data with varying mean and standard deviation. We then measure the similarity between the outputs generated from clean and noisy inputs. Fig.~\ref{fig:similarityOnNoise} presents heatmaps of the batch-averaged similarity scores across different noise configurations. The quantum model exhibits greater robustness to noise, particularly at lower mean values and across a broader standard deviation range. In contrast, the classical model shows a more rapid similarity drop-off, especially when the standard deviation increases. These results highlight the quantum Fourier layer’s resilience and tolerance to input perturbations, indicating its potential advantages in noisy environments.


\section{Related Work}

\textbf{Quantum Scientific Machine Learning.} Quantum Neural Networks(QNNs)~\cite{schuld2014quest,killoran2019continuous,powerofQNN,schuld2021machine}, also referred to as Parametrized Quantum Circuits~\cite{du2020expressive,markidis2023programming}, have been proposed for a range of tasks in scientific computing, including data analysis applications such as classification~\cite{farhi2018classification} and regression~\cite{killoran2019continuous}, as well as the solution of PDEs. This work focuses on the application of QNNs to PDE solving. Two general strategies are typically employed: \emph{(i)} a supervised approach, in which QNNs are trained on datasets generated by classical simulations; and \emph{(ii)} an unsupervised approach, where the governing equations are embedded directly into the loss function, enabling training via automatic differentiation. Methods such as Differentiable Quantum Circuits (DQCs)~\cite{kyriienko2021solving,kyriienko2021generalized} and Quantum Physics-Informed Neural Networks (QPINNs)~\cite{paine2023physics,markidis2022physics} follow this latter paradigm. In the present work, we adopt the supervised strategy, training QNNs on datasets derived from classical numerical solutions.

\noindent \textbf{Fourier Neural Operators for Scientific Computing.} Various QNN architectures have been proposed for solving PDEs. A primary design choice in such architectures is the selection of the variational quantum circuit, which typically consists of parameterized quantum gates combined with data encoding layers~\cite{sim2019expressibility}. In addition to these architectural distinctions, classical neural network models, such as convolutional networks~\cite{kerenidis2019quantum,umeano2023can} and neural operators~\cite{quantumdeepONet} have been adapted to quantum computing frameworks. In this work, we focus on a quantum variant of the neural operator paradigm based on spectral methods, specifically the Fourier Neural Operator (FNO)~\cite{FNO}, which has been successfully used for solving Computational Fluid Dynamics and electromagnetics problems~\cite{costa2023deep,pornthisan2023fast}. Our approach builds on the Quantum Fourier Operator (QFO) framework~\cite{jain2023quantumfouriernetworkssolving}, which incorporates quantum circuits into the FNO architecture by replacing classical Fourier transforms with Quantum Fourier Transforms (QFTs). We extend this method by partitioning the Fourier transform computation across classical and quantum devices to improve scalability and resource efficiency.

\section{Conclusions}
This work presented PH-QFNO, a hybrid quantum-classical architecture for operator learning in scientific machine learning. PH-QFNO extends the QFNO by introducing a partitioning strategy, enabling distributed execution across classical and quantum devices. Inspired by the structure of multidimensional FFTs, the quantum Fourier components are decomposed into parallelizable sub-blocks: each of these is assigned to either classical processors or quantum device. This design allows for tunable degrees of quantum-classical hybridization and scalability across high-dimensional domains.

The method uses unary encoding to map input data into quantum states and trains a parametrized quantum circuit via a variational optimization algorithm. A distributed execution framework based on MPI assigns computational partitions to classical and quantum resources, enabling hybrid emulation using PennyLane’s state vector simulator. The PennyLane implementation is integrated with PyTorch, supporting gradient-based learning.

PH-QFNO was validated on three benchmark problems: Burgers’ equation, the incompressible Navier-Stokes equation, and the compressible Navier-Stokes equation modeling the fluid Kelvin–Helmholtz instability. We showed that PH-QFNO achieves accuracy comparable to classical FNO for Burgers’ equation, even with limited quantum device involvement. For the incompressible Navier-Stokes system, PH-QFNO consistently outperforms classical FNO. Robustness to input noise was evaluated, showing increased stability over classical counterparts.

Future work will focus on extending the current framework to include the impact of quantum device noise, using both realistic noise models available in PennyLane and execution on actual quantum hardware. While variational training on emulated QPUs remains feasible, communication latency between classical and quantum components is expected to be a dominant performance bottleneck in real deployments due to the iterative nature of gradient evaluation and circuit updates. Algorithms, such as PH-QFNO, require high integration of classical and quantum devices and avoid communication across the internet.

Algorithmically, the current implementation is purely data-driven, with training samples obtained from finite difference and finite volume classical simulations. A future extension will incorporate physics-based loss functions that penalize the residual of the governing PDEs, following the principles of DQC~\cite{kyriienko2021solving} and QPINNs~\cite{markidis2022physics}. The inclusion of a physics-informed loss function part will improve generalization and enforce physical conservation laws.

\section*{Acknowledgments} 
\noindent Partial financial support from ICSC - “National Research Centre in High Performance Computing, Big Data and Quantum Computing”, funded by European Union – NextGenerationEU, is gratefully acknowledged. S.M. acknowledges support from a Center of Excellence in Exascale CFD (CEEC) grant No 101093393 funded by the European Union via the European High Performance Computing Joint Undertaking (EuroHPC JU) and Sweden, Germany, Spain, Greece and Denmark.


\end{document}